\newif\ifconfver
\confvertrue        

\ifconfver
\documentclass[10pt,twocolumn,twoside]{IEEEtran}
\else
\documentclass[11pt,draftcls,onecolumn]{IEEEtran}
\fi

\usepackage{makecell}
\usepackage{bbding}
\usepackage{color,graphicx}
\usepackage{float}
\usepackage{multirow,amsmath,epsfig,amsfonts,amssymb,psfig,graphics,psfrag,theorem,calc,url,bm,cite}
\usepackage{stfloats,hyperref}
\usepackage{algorithm,algorithmic}
\usepackage{diagbox} 
\usepackage{hhline}
\usepackage{subfigure}
\usepackage{theorem}
\usepackage{comment}
\usepackage{pifont}
\usepackage{tikz}
\usepackage{fancyhdr}
\usetikzlibrary{quantikz2}

\usepackage{mathtools}

\makeatletter
\def\multilimits@{\bgroup
	\Let@
	\restore@math@cr
	\default@tag
	\baselineskip\fontdimen10 \scriptfont\tw@
	\advance\baselineskip\fontdimen12 \scriptfont\tw@
	\lineskip\thr@@\fontdimen8 \scriptfont\thr@@
	\lineskiplimit\lineskip
	\vbox\bgroup\ialign\bgroup\hfil$\m@th\scriptstyle{##}$\hfil\crcr}
\def\Sb{_\multilimits@}
\def\endSb{\crcr\egroup\egroup\egroup}
\makeatother

{\begin{list}{}%
		{\setlength{\rightmargin}{0pc}%
			\setlength{\leftmargin}{1pc} }} %
	{\end{list}}

\newlength{\twidth}
\ifconfver
\else
\fi

\definecolor{orange}{RGB}{255,107,0}

\newtheorem{Theorem}{Theorem}

\theorembodyfont{\rmfamily}

{\begin{list}{}{
			\settowidth{\labelwidth}{\mbox{\textnormal{#1}}}%
			\setlength{\leftmargin}{\labelwidth+\labelsep}}}%
	{\end{list}}

\newcommand\bI{\ensuremath{{\bm I}}}

\newcommand\bM{\ensuremath{{\bm M}}}

\newcommand\bW{\ensuremath{{\bm W}}}
\newcommand\bX{\ensuremath{{\bm X}}}

\newcommand\bZ{\ensuremath{{\bm Z}}}

\definecolor{orange}{RGB}{255,107,0}

\ifconfver
\author{Chia-Hsiang Lin,~\IEEEmembership{Member,~IEEE},
Tzu-Hsuan Lin,~\IEEEmembership{Graduate Student Member,~IEEE},
\\and Jocelyn Chanussot,~\IEEEmembership{Fellow,~IEEE}}
\title{Quantum Information-Empowered Graph Neural 

~~~~Network for Hyperspectral Change Detection%
    \thanks{This study was supported in part by the Emerging Young Scholar Program (namely, the 2030 Cross-Generation Young Scholars Program) of National Science and Technology Council (NSTC), Taiwan, under Grant NSTC 113-2628-E-006-003, and in part by the Ph.D. Students Study Abroad Program of NSTC, Taiwan, under Grant NSTC 113-2917-I-006-012.
   We thank the National Center for Theoretical Sciences (NCTS) and the National Center for High-performance Computing (NCHC) for providing the computing resources.}
	\thanks{\textit{(Corresponding author: Chia-Hsiang Lin.)}}
	\thanks{Chia-Hsiang Lin is with the Department of Electrical Engineering, and with the Miin Wu School of Computing, National Cheng Kung University, Tainan, Taiwan (R.O.C.) (e-mail: chiahsiang.steven.lin@gmail.com).}
    \thanks{Tzu-Hsuan Lin is with the Institute of Computer
and Communication Engineering, Department of Electrical Engineering, National Cheng Kung University, Tainan, Taiwan (R.O.C.) (e-mail: q38091518@gs.ncku.edu.tw).}	

\thanks{Jocelyn Chanussot is with Inria, CNRS, Grenoble INP, LJK, Université Grenoble Alpes, 38000 Grenoble, France (e-mail: jocelyn.chanussot@inria.fr).}
}

\fancypagestyle{IEEEtitlepagestyle}{
	\fancyhf{} 
	\fancyhead[LE,RO]{\thepage} 
	\fancyfoot[C]{\scriptsize © 2024 IEEE. Personal use of this material is permitted. Permission from IEEE must be obtained for all other uses, in any current or future media, including reprinting/republishing this material for advertising or promotional purposes, creating new collective works, for resale or redistribution to servers or lists, or reuse of any copyrighted component of this work in other works.}
	
}

\else

\fi
\begin{document}
	
	\bibliographystyle{IEEEtran}
	\maketitle
	\ifconfver \else \vspace{-0.5cm}\fi

	\begin{abstract}

        Change detection (CD) is a critical remote sensing technique for identifying changes in the Earth's surface over time.
        The outstanding substance identifiability of hyperspectral images (HSIs) has significantly enhanced the detection accuracy, making hyperspectral change detection (HCD) an essential technology.
        The detection accuracy can be further upgraded by leveraging the graph structure of HSIs, motivating us to adopt the graph neural networks (GNNs) in solving HCD.
        For the first time, this work introduces quantum deep network (QUEEN) into HCD.
        Unlike GNN and CNN, both extracting the affine-computing features, QUEEN provides fundamentally different unitary-computing features.
        We demonstrate that through the unitary feature extraction procedure, QUEEN provides radically new information for deciding whether there is a change or not.
        Hierarchically, a graph feature learning (GFL) module exploits the graph structure of the bitemporal HSIs at the superpixel level, while a quantum feature learning (QFL) module learns the quantum features at the pixel level, as a complementary to GFL by preserving pixel-level detailed spatial information not retained in the superpixels.
        In the final classification stage, a quantum classifier is designed to cooperate with a traditional fully connected classifier.
        The superior HCD performance of the proposed QUEEN-empowered GNN (i.e., QUEEN-$\mathcal{G}$) will be experimentally demonstrated on real hyperspectral datasets.
        
		
	\end{abstract}

	\ifconfver \else \vspace{-0.0cm}\fi
	
	\ifconfver \else \vspace{-0.5cm}\fi
	
	\ifconfver \else  \fi

\section{Introduction} \label{sec: introduction}

Change detection (CD) aims to detect the changed areas of the same region between two different dates using remote sensing images, making this technique crucial for monitoring the earth's surface.
CD has had a profound impact on various substantial applications such as agriculture investigation, disaster management, urban planning, and land cover/use monitoring \cite{abd2011land, bovolo2007split, yang2003urban}.
The evolution of hyperspectral imaging (HSI) technology has revolutionized various remote sensing applications \cite{bioucas2013hyperspectral,li2024learning,lin2021all, lin2024signal,du2012fusion}.
Traditional CD datasets mainly include very high resolution (VHR) satellite images, synthetic aperture radar (SAR) images, and multispectral images (MSIs) \cite{falco2012change, rignot1993change, zheng2019unsupervised}.
Leveraging the rich spectral information, HSI data further enables outstanding substance identifiability \cite{lin2015identifiability, lin2018maximum}, thereby significantly enhancing the detection accuracy.
In light of this advantage, hyperspectral change detection (HCD) has gained significant traction in recent years.

Conventional CD methods can be divided into three categories: arithmetic-based, transformation-based, and classification-based.
Arithmetic-based methods primarily utilize mathematical operations to detect the changed pixels between the bitemporal HSIs.
For example, change vector analysis (CVA) detects the changes in multispectral images by analyzing the change vectors, with the possibility of change being the magnitude of the change vectors \cite{bovolo2006theoretical}.
For transformation-based methods, bitemporal images are transformed into specialized feature domains.
This process aims to suppress the unchanged pixels while highlighting the changed pixels \cite{wu2013slow,tewkesbury2015critical}.
For instance, multivariate alteration detection (MAD) leverages a transformation based on correlation analysis to perform a pixel-wise change detection of bitemporal images \cite{nielsen1998multivariate}.
Classification-based methods treat the CD task as a classification problem and employ classifiers to categorize the pixels into changed and unchanged classes.
A method called support vector machines (SVMs) integrates multiple SVM classifiers into a framework by combining the classification output of the classifiers \cite{nemmour2006multiple}.
Besides the three main conventional categories, the CD problem has also been formulated as a convex optimization (CO) problem and solved by the convex optimizer alternating direction method of multipliers (ADMM) with convergence guaranteed \cite{zheng2019unsupervised}. 
Despite the contributions of traditional change detection methods, their reliance on manually designed features becomes a critical limitation in HCD analysis.

With the development of deep learning (DL), convolutional neural networks (CNNs) have demonstrated remarkable success across diverse domains, revolutionizing numerous fields of study and application \cite{guo2016deep, deng2014deep}.
This widespread adoption has naturally extended to the field of remote sensing, where CNNs are increasingly being leveraged to address complex challenges in HSI processing \cite{lin2024qrcode}.
In recent years, CNN-based approaches for HCD have drawn attention due to their ability to capture spectral-spatial features and overcome limitations existing in traditional methods.
A single-stream CNN framework has been proposed to detect the bitemporal changed areas using remote sensing images \cite{dong2018local}. 
We have to mention GETNET, an end-to-end 2-D CNN architecture, which employs a mixed-affinity matrix to extract cross-channel gradient information \cite{wang2018getnet}.
Moreover, Bilinear Convolutional Neural Network (BCNN), a siamese-structured architecture, captures inter-temporal relationships in bitemporal images by employing a symmetric CNN and fusing the output feature maps \cite{lin2019multispectral}.
In multilevel encoder-decoder attention network (ML-EDAN) \cite{qu2021multilevel}, a two-stream encoder-decoder framework with a contextual-information-guided attention module is developed to obtain the multilevel hierarchical features, and the long short-term memory (LSTM) subnetwork is incorporated to analyze temporal correlations.
In addition to conventional CNNs, transformers, a novel neural network architecture that leverages self-attention mechanisms to process sequential data in parallel, offering a superior ability to capture long-range dependencies and contextual information, has successfully been adopted to solve challenging hyperspectral problems \cite{tang2024transformer, young2023cidar}.
The concept of self-attention mechanisms has recently been utilized to design neural networks for HCD.
CSANet employs a Siamese 2-D convolutional neural network architecture incorporating attention mechanisms to effectively extract the spatial-spectral-temporal features from bitemporal HSI pairs \cite{song2022csanet}.
Furthermore, the spectral–spatial-temporal transformer (SST-Former) integrates spectral, spatial, and temporal transformers in a sequential manner, employing position encoding, spectral and spatial transformer encoders, class tokens, and a temporal transformer to process multi-dimensional HSI data \cite{wang2022spectral}.
In multiscale diff-changed feature fusion network (MSDFFN) \cite{luo2023multiscale}, a temporal feature encoder–decoder (TFED) subnetwork with a bidirectional diffchanged feature representation (BDFR) module and a multiscale attention fusion (MSAF) module is proposed to extract multiscale features.
However, Transformer-based models typically have a complex network structure, which leads to the potential overfitting on limited labeled samples when solving the HCD problem.
Therefore, a gated spectral–spatial–temporal attention network with spectral similarity filtering (HyGSTAN) \cite{yu2024hyperspectral} employs a spectral similarity filtering module (SSFM) to reduce spectral redundancy within patches, followed by a lightweight gated spectral-spatial attention module (GS2AM) for extracting long-range spectral dependencies.
Finally, HyGSTAN integrates temporal information through a gated spectral-spatial-temporal attention module (GS2TAM), ultimately reconstructing change information through an inverse spectrum merging operation.
Although CNN-based methods have achieved promising performance, they have not fully exploited the characteristics of HSIs, such as the graph structure, to attain better results.

Recently, graph neural networks (GNN) have been adopted to tackle hyperspectral remote sensing tasks due to the ability to learn the long-range information of the HSIs \cite{hong2020graph, yu2022unsupervised, lin2023gnn}.
GNNs have also been designed for the CD problem; for example, an unsupervised structural relationship graph representation learning framework is proposed for multimodal change detection (SRGRL-CD) \cite{chen2022unsupervised}.
Moreover, dual-branch difference amplification graph convolutional network ($\textnormal{D}^2$AGCN) innovatively integrates a dual-branch structure with a difference amplification module (DAM), leveraging graph convolutional networks (GCNs) to extract and amplify distinctive features from bitemporal HSIs, thereby enhancing the discrimination between changed and unchanged areas even with limited training samples \cite{qu2021dual}.
By introducing the concept of graph attention network (GAT) \cite{velivckovic2017graph},  a two-branch framework with a novel temporal-spatial joint graph attention (TSJGAT) module in superpixel-level branch and a CNN-based pixel-level branch is proposed in complement strategy dual-branch framework (CSDBF), enabling parallel extraction and complementary fusion of multi-scale features from bitemporal HSIs, thus enhancing discrimination of changed regions while mitigating uncertainties introduced by superpixel segmentation \cite{wang2022csdbf}.
To address the issue that graph structure constructed by superpixels ignores the multi-order difference information among graph nodes and the local difference information within superpixels, an efficient multi-order GCN (MGCN) with a channel attention module (CAM) is proposed, iteratively refining neighborhood feature representations and enhancing bitemporal difference features \cite{zhang2023multi}.
Synergizing the strengths of GNN and transformer, a dual-branch local information-enhanced graph-transformer (D-LIEG) network combing cascaded LIEG blocks with a novel graph-transformer is proposed, effectively extracting and integrating local-global spectral-spatial features from bitemporal HSIs, thus enabling robust change recognition even with limited training samples while preserving spectral information and accommodating diverse change areas \cite{dong2023local}.
We have to mention a convex deep learning algorithm for HCD, termed CODE-HCD \cite{lin2023hyperspectral}.
CODE-HCD is designed based on a framework called CODE \cite{lin2021admm}, which combines the advantages of CO and DL, alleviating the mathematical burden in the data-fitting design in the CO part and tolerating the small-data situation for training the GNN CD network in the DL part.
Although GNN-based methods have exploited the graph structure of HSIs, these conventional DL approaches, including CNN-based and GNN-based ones, focus on learning affine-mapping features, which may limit their ability to capture more diverse information.
We believe that introducing radically new information sources is critical for any decision-making procedure (e.g., the final detection procedure in HCD).
This motivates us to introduce the novel quantum unitary-computing information for obtaining more effective HCD solutions, though integrating the quantum information with the traditional feature information has posed a great challenge (to be solved in this work).

In recent years, quantum computing has emerged as a promising field, garnering significant attention from researchers.
Quantum neural networks (QNNs), designed based on parameterized quantum circuits, have been developed to address complex problems and have demonstrated exceptional performance in diverse domains, including drug response prediction, health detection, and image classification \cite{sagingalieva2023hybrid, qu2023iomt, fan2023hybrid}.
These problems are mostly classification tasks, and those related to images are mainly RGB image problems.
However, due to the limitations of modern quantum computers, complementary techniques are essential to achieve advanced quantum computing capabilities for more sophisticated applications \cite{marvian2022restrictions}.
In 2023, an innovative quantum deep network (QUEEN) for addressing the challenging hyperspectral inpainting problem was developed as the first quantum algorithm for image restoration \cite{lin2023hyperqueen}.
By incorporating CNN to assist QNN, it can address the constraints imposed by limited quantum bits (qubits) resources and the barren plateaus (BP) effect to restore HSIs.
The QUEEN theory demonstrates the efficacy of quantum unitary-computing features extracted by quantum networks in addressing diverse imaging inverse problems.
Very recently, QUEEN has been further applied to critical remote sensing missions, such as the challenging super-resolution task and the highly challenging multispectral unmixing (MU) problem \cite{hsu2024hyperqueen, lin2024prime}.

\begin{figure*}[t]
    \centering
    \includegraphics[width=0.99\linewidth]{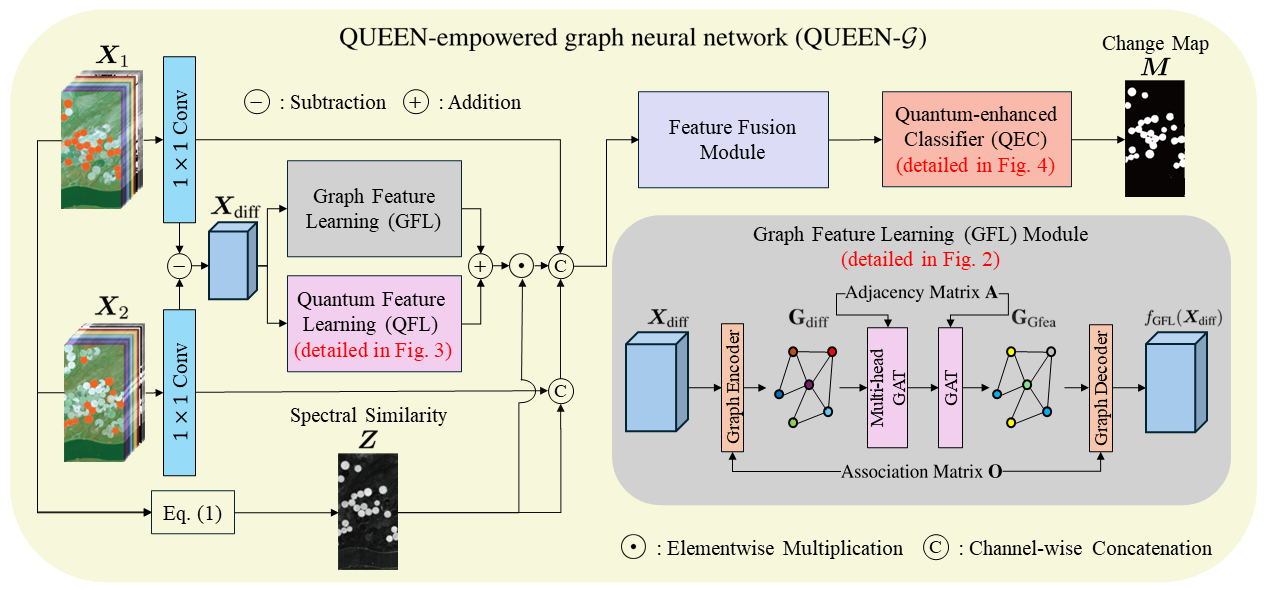}
    \caption{Overall network framework of the proposed QUEEN-$\mathcal{G}$. 
    The bitemporal HSIs are first dimension-reduced by a $1 \times 1$ convolutional layer.
    Subsequently, the graph feature learning (GFL) and quantum feature learning (QFL) modules, further detailed in Figs. \ref{fig:GFL} and \ref{fig:QFL}, are adopted to learn the graph-structured and quantum unitary-computing features at the superpixel level and pixel level, respectively.
    The extracted quantum feature and graph feature are then fused, and weighted by the spectral similarity matrix $\bZ$ (obtained from $\bX_1$ and $\bX_2$).
    Then, the extracted feature map, the dimension-reduced input HSIs, and the spectral similarity map are further fused by a feature fusion module.
    Finally, the quantum-enhanced classifier (QEC), detailed in Fig. \ref{fig:QEC}, is employed to detect the changed areas.
    With the additional quantum branch, new feature information yields upgraded decision-making in the final classification/detection.
    }\label{fig:architecture}
\end{figure*}

In this paper, we propose a QUEEN-empowered graph neural network (QUEEN-$\mathcal{G}$) for the HCD problem, integrating GNN and QNN to extract features at the superpixel level and pixel level, respectively.
We remark that the design of QUEEN-based architecture is challenging because some powerful traditional architectures cannot be implemented under the quantum framework; for example, the quantum no-cloning theorem limits the possibility of implementing the quantum version of the powerful residual network (ResNet) \cite{he2016deep}.
Compared to GNN, which extracts affine mapping features, QUEEN provides unitary-computing
capabilities during feature extractions \cite{lin2023hyperqueen}, thus obtaining features from a radically new perspective for decision-making in the final detection.
Moreover, the proposed algorithm incorporates a quantum-enhanced classifier (QEC) to improve the detection of changed areas.
We propose a quantum architecture with full expressibility (FE) for the QUEEN-based modules used in QUEEN-$\mathcal{G}$, meaning that the quantum network in the quantum modules can generate any possible quantum state or perform any unitary transform \cite{lin2023hyperqueen}.
This property allows the QFL and QEC module to effectively extract useful pixel-level entangled quantum features using the quantum unitary operators.
The superpixel-level affine-mapping features are learned by a graph feature learning (GFL) module constructed based on \cite{lin2023hyperspectral}, leveraging the tendency of changed pixels to form small regions.
At the same time, a quantum feature learning (QFL) module extracts pixel-level entangled unitary-computing features, serving as a complementary mechanism to the GFL by preserving detailed spatial information not retained in the superpixels.
In light of the outstanding classification ability of QNN that has been proven in the existing literature, the final classification stage employs the quantum-enhanced classifier, QEC, that fuses a quantum classifier with a traditional fully connected layer, utilizing a customized loss function to enhance detection performance.
For the first time, QUEEN has been introduced for solving the highly challenging HCD problem.
The main contributions of this article are summarized
as follows:
\begin{enumerate}
  \item 
  In this paper, we propose a QUEEN-based quantum HCD algorithm, termed as QUEEN-$\mathcal{G}$, by incorporating GNN and QNN, where the GFL and QFL modules extract the global superpixel-level and local pixel-level features, respectively.
  The entangled unitary-computing QNN features are fused with the traditional affine-computing GNN features, providing radically new quantum information for better decision-making in the final detection stage of QUEEN-$\mathcal{G}$.

  \item 
  The QUEEN has the mathematically provable FE, so that the quantum network in the QUEEN-$\mathcal{G}$ can generate any possible quantum state or perform any valid quantum transform.
  This property allows the QFL module to effectively extract useful pixel-level entangled quantum features using the quantum unitary operators.
  QFL, serving as a complementary role to the superpixel-level GFL module, is designed to work under the pixel-level mechanism for preserving the detailed spatial information not retained in the superpixels.
  As will be seen, the new quantum information yields the state-of-the-art HCD performances, as evidenced by the ablation study (cf. Section \ref{sec:ablation}).
  
  \item 
  The QEC, which integrates a quantum classifier with a traditional fully connected layer, is constructed for the final classification stage (decision-making stage).
  Meanwhile, a customized loss function is proposed to individually enhance the classification ability of the quantum classifier and the traditional classifier, thereby improving the detection performance.
  The effectiveness of the QEC is also proved by the ablation study in Section \ref{sec:ablation}.
  
  \item
  The proposed QUEEN-$\mathcal{G}$ significantly surpasses existing benchmark HCD methods, achieving state-of-the-art performance on real hyperspectral datasets.
  As detailed in Section \ref{sec:experiment}, both the quantitative analysis and the qualitative analysis demonstrate the superiority of QUEEN-$\mathcal{G}$ in accurately detecting the changed areas.
\end{enumerate}

The remaining parts of the paper are organized as follows.
In Section \ref{sec:method}, the design of the proposed QUEEN-$\mathcal{G}$ for the HCD problem is illustrated.
In Section \ref{sec:experiment}, experiments on real hyperspectral datasets of the comparisons with benchmark HCD methods and the analysis of the proposed algorithm are summarized and discussed.
Finally, conclusions are drawn in Section \ref{sec:conclusion}.

\section{Proposed Method} \label{sec:method}

\subsection{Network Architecture}\label{sec:architecture}

The overall network architecture of the proposed QUEEN-empowered graph neural network, QUEEN-$\mathcal{G}$, for the HCD problem is graphically illustrated in Fig. \ref{fig:architecture}.
The proposed method can be broadly separated as a two-stage process.
First, features are independently extracted at the superpixel level using a GNN-based module and at the pixel level using a QNN-based module.
The two branches provide fundamentally different features, i.e., the GNN-based branch extracts affine-mapping features while the QNN-based branch captures entangled unitary-computing features, thus providing radically new information for better decision-making in the final detection.
These features are subsequently fused and passed through a final classifier combining traditional fully connected and QUEEN to detect areas of change.
Next, we will describe the proposed network architecture in detail.

Given the bitemporal HSIs $\bX_1, \bX_2\in\mathbb{R}^{H\times W\times C}$ acquired at different dates and covering the same spatial area, as the input of QUEEN-$\mathcal{G}$, where $H$ and $W$ denote the height and the width of the region of interest (ROI), respectively, and $C$ represents the number of spectral bands.
In the first layer of the network, a $1 \times 1$ convolutional layer is adopted to project the spectral dimension of $\bX_1$ and $\bX_2$ onto a lower-dimensional subspace due to the high number of spectral bands in HSIs, which can be written as
\begin{equation*}
\label{eq:DR}
    \widetilde{\bX}_i=f_\textnormal{Conv}^{1\times 1}(\bX_i), ~~i=1,2,
\end{equation*}
where $\widetilde{\bX}_i\in\mathbb{R}^{H\times W\times 64}$, effectively reducing spectral redundancy and minimizing the number of parameters in the network.
After obtaining the dimension-reduced bitemporal images, the differential feature map, i.e., the subtraction of $\bX_1$ and $\bX_2$, defined as
\begin{equation*}
\label{eq:sub}
    \bX_\textnormal{diff}=\widetilde{\bX}_1-\widetilde{\bX}_2,
\end{equation*}
is computed, and its features are subsequently learned by the network.
Next, the GFL module extracts the superpixel-level affine-mapping features using graph convolutional layers.
At the same time, the QFL module, designed based on QUEEN, learns the pixel-level quantum unitary-computing features as a complementary role to the GFL module, preserving the detailed spatial information not retained in the superpixels.
The two feature maps generated by the GFL and QFL modules are then fused with addition.
Considering the outstanding substance identifiability of HSIs, the spectral changes of the bitemporal HSIs imply material changes.
This motivates us to introduce the spectral angle mapper (SAM) to generate the spectral similarity map $\bZ$ of the bitemporal HSIs, whose $(i,j)$th element $\bZ_{i, j}$ can be mathematically written as
\begin{equation}
    \bZ_{i, j} = \arccos \left(\frac{\sum_{k=1}^{C} \bX_1(i, j, k) \bX_2(i, j, k)}{\sqrt{\sum_{k=1}^{C} \bX_1(i, j, k)^2} \sqrt{\sum_{k=1}^{C} \bX_2(i, j, k)^2}}\right),
\end{equation}
where $\bX_1(i, j, k)$ and $\bX_2(i, j, k)$ denote the $(i, j, k)$th element of $\bX_1$ and $\bX_2$, respectively, as the weight along the spatial dimension of the fused feature map.
%
The larger the value of SAM, the more different the spectral shapes are (indicating a higher chance to be a changed pixel).
Therefore, by multiplying the weight $\bZ$ along the spatial dimension, we can highlight the pixels that have a higher probability of being a changed pixel based on the characteristic of HSI.
Therefore, the weighted feature map can be expressed as
\begin{equation*}
\label{eq:weight}
    \bX_\textnormal{fea}=[f_\textnormal{GFL}(\bX_\textnormal{diff})+f_\textnormal{QFL}(\bX_\textnormal{diff})] \odot \bZ,
\end{equation*}
where $\odot$ denotes elementwise multiplication.
In the next stage, the weighted feature map, the dimension-reduced bitemporal images, and the spectral similarity map are fed into the feature fusion module.
The feature fusion module comprises two feature fusion layers (FF), each consisting of a $3 \times 3$ convolutional layer followed by a PReLU activation function and a batch normalization layer.
The feature fusion module and the fused feature map $\bX_\textnormal{fuse}$ can be explicitly written as
\begin{align*}
    f_\textnormal{FF}(\cdot)&=\textnormal{BN}(\sigma(f_\textnormal{conv}^{3 \times 3}(\cdot))), \\    \bX_\textnormal{fuse}&=f_\textnormal{FF}(f_\textnormal{FF}(\bX_\textnormal{fea}\|\widetilde{\bX}_1\|\widetilde{\bX}_2\|\bZ)),
\end{align*}
where $\sigma$, BN, and $\|$ denote the PReLU activation, the batch normalization layer, and the channel-wise concatenation, respectively.
In the final classification stage, the output change map $\bM$ is generated by QEC, which is a classifier incorporating the quantum classifier and the traditional fully connected layer (c.f. Fig. \ref{fig:QEC}), and can be expressed as 
\begin{equation*}
\label{eq:classifier}
    \bM=f_\textnormal{QEC}(\bX_\textnormal{fuse}),
\end{equation*}
where $f_\textnormal{QEC}$ will be explicitly designed in Section \ref{sec: Qclass}.

\subsection{Superpixel-level Feature Learning using GFL}\label{sec:GNN}

\begin{figure}[t]
    \centering
    \includegraphics[width=1\linewidth]{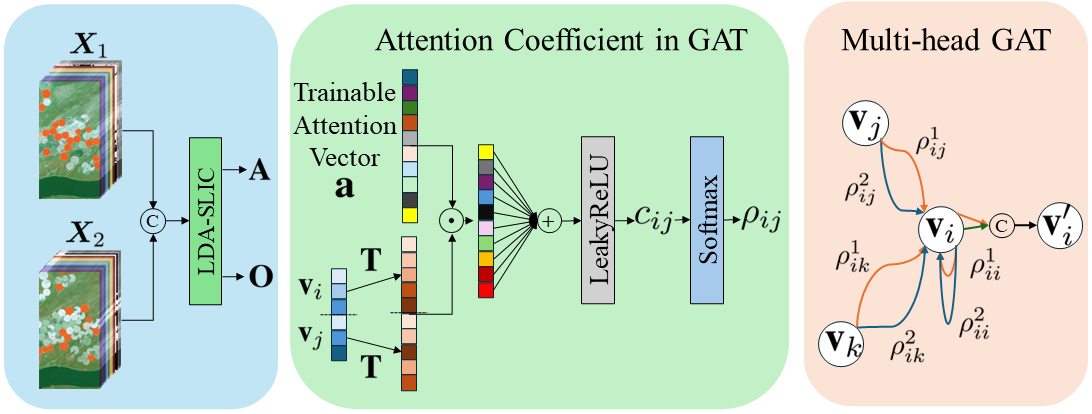}
    \caption{Detailed architecture of the GFL module, which is outlined in Fig. \ref{fig:architecture}.
    The GFL module begins by converting the pixel-level feature map into a graph using a graph encoder.
    It then extracts graph-structured features at the superpixel level through a multi-head GAT layer followed by a single-head GAT layer.
    Finally, the feature graph is transformed back to the pixel level via a graph decoder.
    The block at the left describes the generation of the adjacency matrix $\textbf{A}$ and the association matrix ${\textbf{O}}$.
    The block in the middle details the process to compute the attention coefficient $\rho_{i j}$ in the GAT layers.
    The block at the right illustrates the information propagation of the nodes in the multi-head GAT layer.
}\label{fig:GFL}
\end{figure}

The GFL module is constructed by GNN layers to extract the graph-structured superpixel-level features as illustrated in Fig. \ref{fig:GFL}.
Since the changed areas tend to form small regions, learning the embedded information at the superpixel level can be crucial \cite{lin2023hyperspectral}.
The GAT layers, which propagate information from the first-order neighborhood to a node, are employed to enable each superpixel to learn from its surrounding superpixels.
The GFL module consists of a multi-head GAT layer and a single-head GAT layer.
Before feeding the feature map into the GFL module, we need to convert $\bX_\textnormal{diff}$ into a graph structure to meet the requirements of the GAT layers.

To transform the differential feature map into superpixel-level to form the desired graph, the input bitemporal HSIs are firstly concatenated along the channel dimension, i.e., $\bX_1\|\bX_2$.
The concatenated HSI is then dimension-reduced using linear discriminant analysis (LDA) \cite{xanthopoulos2013linear}, and finally fed into the simple linear iterative clustering (SLIC) method \cite{achanta2012slic} to obtain the association matrix $\textbf{O}$ and the adjacency matrix $\textbf{A}$ for constructing the graph.

The graph of the differential feature map, i.e., $\textbf{G}_\textnormal{diff}\triangleq (\textbf{V}_\textnormal{diff}, \textbf{E}_\textnormal{diff})$ with $\textbf{V}_\textnormal{diff}$ and $\textbf{E}_\textnormal{diff}$ denoting the nodes and edges, respectively, is then constructed by a graph encoder as
\begin{align*}
\textbf{O}_{i, j}&=\left\{\begin{array}{ll}
1, & \text { if } \textbf{x}_i \in \mathcal{S}_j, \\
0, & \text { otherwise, }\\
\end{array} \right.\\
{\widetilde{\textbf{O}}}_{i, j}&=\frac{\textbf{O}_{i, j}}{\sum_{l=1}^HW \textbf{O}_{l, j}}, \\
\textbf{V}_\textnormal{diff}&=f_\textnormal{Encoder}(\textbf{X}_\textnormal{diff}, \textbf{O})={\widetilde{\textbf{O}}}^T \textbf{X}_\textnormal{diff}^T,
\end{align*}
where $\mathcal{S}_i$ denotes the $i$th superpixel, $\textbf{X}_\textnormal{diff}\in\mathbb{R}^{64\times HW}$ stands for the 2-D version of the 3-D tensor $\bX_\textnormal{diff}\in\mathbb{R}^{H\times W \times 64}$, and $\textbf{x}_i$ represents the $i$th column of $\textbf{X}_\textnormal{diff}$.

The GAT layers propagate the information of the first-order neighborhood (i.e., the surrounding superpixel) to the node (i.e., the superpixel).
To compute the aggregate coefficient between two nodes $\textbf{v}_i$ and $\textbf{v}_j$, where $\textbf{v}^T_k$ denotes the $k$th row of $\textbf{V}_\textnormal{diff}$, the two nodes are firstly transformed by a learnable transformation $\textbf{T}$ to a high-level feature space.
Next, a trainable attention vector $\textbf{a}$ is applied to preserve the crucial elements of the transformed nodes, followed by a LeakyReLU activation with a slope set as 0.2 to obtain an attention coefficient $c_{ij}$.
Subsequently, the attention coefficients $\{\rho_{i j}\}$ of the first-order neighborhood are then normalized using the Softmax function, which can be detailed as
\begin{align*}
    c_{i j}&=\text{LeakyReLU}(\textbf{a}^T(\textbf{T}\textbf{v}_i\| \textbf{T}\textbf{v}_j)),\\
    \rho_{i j}&=\text{Softmax}_j\left(c_{i j}\right)=\frac{\exp \left(c_{i j}\right)}{\sum_{k \in \mathcal{N}_i} \exp \left(c_{i k}\right)},
\end{align*}
where $ \mathcal{N}_i$ denotes the first-order neighborhood of $i$.
At the first layer of GFL, a 2-head GAT layer is adopted to stabilize the learning process \cite{velivckovic2017graph} and fully exploit the embedded information, which can be expressed as
\begin{equation*}
    \textbf{v}'_i=\|_{n=1}^2\text{ELU}\left(\sum_{j \in \mathcal{N}_i} \rho_{i j}^n \textbf{T}_1^n \textbf{v}_j\right),
\end{equation*}
where the subscript $n$ denotes the $n$th head, and the exponential linear unit (ELU) function is selected as the activation.

Subsequently, a single GAT layer is employed to fuse the multi-head features and to learn the information from the second-order neighborhood, producing the final feature graph $\textbf{V}_\textnormal{Gfea}$, whose $i$th column $\textbf{v}_{\textnormal{Gfea}_i}$ can be expressed as
\begin{equation*}\label{eq:2ndGAT}
    \textbf{v}_{\textnormal{Gfea}_i}=\text{ELU}\left(\sum_{j \in \mathcal{N}_i} \nu_{i j} \textbf{T}_2 \textbf{v}'_j\right),
\end{equation*}
where $\nu_{i j}$ is the attention coefficient for the second layer.
After finishing learning features at the superpixel level, the final feature graph $\textbf{V}_\textnormal{Gfea}$ needs to be converted back to a pixel-level feature map, which can be achieved by the graph decoder, i.e.,
\begin{equation*}
    f_\textnormal{GFL}(\bX_\textnormal{diff})=f_\textnormal{Decoder}(\textbf{V}_\textnormal{Gfea}, \textbf{O})=\textbf{O} \textbf{V}_\textnormal{Gfea}.
\end{equation*}
At this point, we have successfully exploited the graph structure of the bitemporal HSI by extracting the superpixel-level feature using the GFL module.
The overall GFL function has been summarized in Fig. \ref{fig:GFL}.

\subsection{Pixel-level Feature Learning using QFL}\label{sec:QNN}

\begin{figure}[t]
    \centering
    \includegraphics[width=1\linewidth]{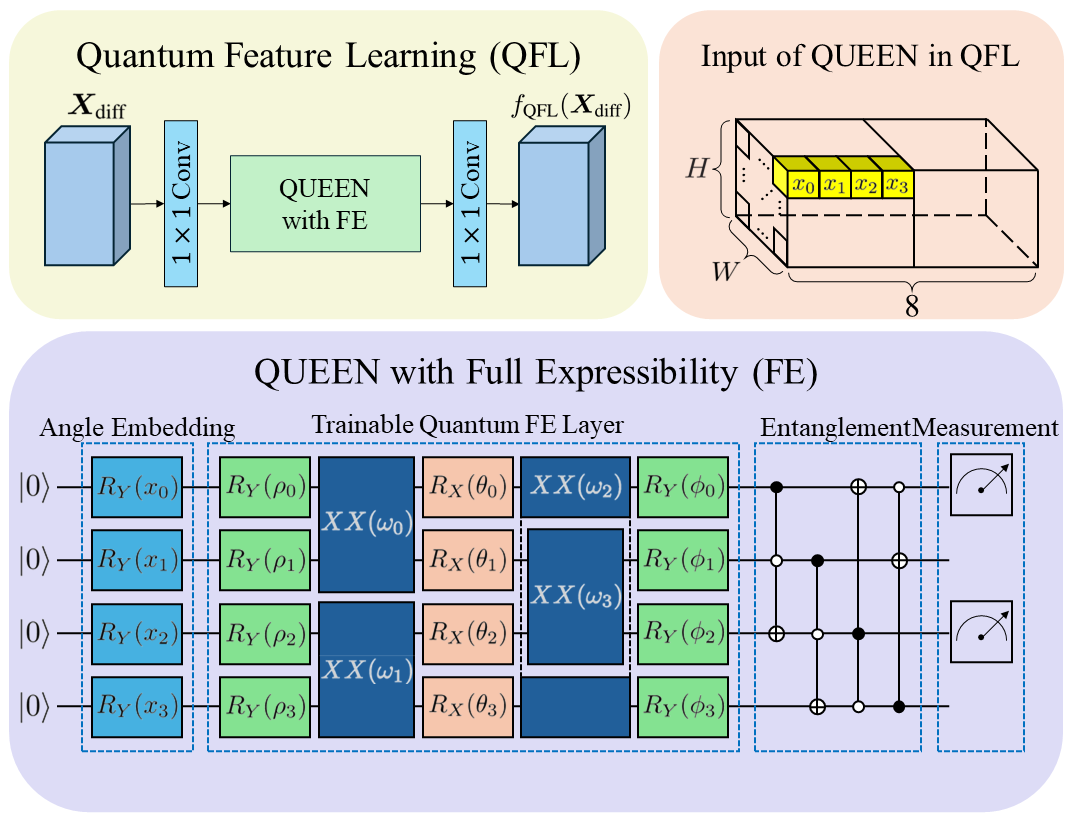}
    \caption{Detailed architectures of the QFL module, and of the QUEEN (together with its input dimension).
    The block at the top left is the QFL module, which is composed of a $1 \times 1$ convolutional layer, a QUEEN with FE to extract quantum unitary-computing features at the pixel level, and a $1 \times 1$ convolutional layer.
    The block at the bottom is the QUEEN with mathematically provable FE that can generate any possible quantum state or implement any valid quantum function.
    The block at the top right illustrates how to input the feature map into the QUEEN for learning the quantum information.
    }\label{fig:QFL}
\end{figure}

\begin{table}[t]
\scriptsize
\setlength{\tabcolsep}{5.8pt} 
\caption{Symbols and mathematical definitions for the quantum gates used in the QUEEN architecture (cf. Figure \ref{fig:QFL}), all corresponding to some unitary operators, where $\theta$ represents the learnable parameters.  
For conciseness, we use $\delta$ and $\gamma$ to denote $\cos(\frac{\theta}{2})$ and $\sin(\frac{\theta}{2})$, respectively.
Also, DIAG($A$, $B$, $C$) is a block-diagonal matrix with diagonal blocks $A$, $B$ and $C$.
Finally, $\bI_n$ denotes the $n\times n$ identity matrix.}\label{tab:common_qu_gate}
\begin{center}
\renewcommand\arraystretch{1.3}
\begin{tabular}{|c c c|} 
 \hline
 \rule{0pt}{2ex}
 Quantum Gate & Symbol & Unitary Operator 
 \rule{0pt}{2ex}
 \\
 \hline
 \rule{0pt}{4ex}
 Rotation X
 &
 \begin{tikzcd}
    \qw & \gate{R_{X}(\theta)} & \qw
 \end{tikzcd}
 &
 $\begin{pmatrix}
    \delta & -i \gamma \\
    -i \gamma & \delta
\end{pmatrix}$
 \rule{0pt}{4ex}
 \\ 
 \hline
 \rule{0pt}{4ex}
 Rotation Y
 &
 \begin{tikzcd}
    \qw & \gate{R_{Y}(\theta)} & \qw 
 \end{tikzcd}
 & 
 $\begin{pmatrix}
    \delta & - \gamma \\
    \gamma & \delta
\end{pmatrix}$
 \rule{0pt}{4ex}
\\
 \hline
 \rule{0pt}{4ex}
 \rule{0pt}{6.5ex}
 Ising XX
&
\begin{quantikz}
    \qw & \gate{XX(\theta)} & \qw
\end{quantikz}
&
$\begin{pmatrix}
    \delta & 0 & 0 & -i\gamma\\
    0 & \delta & -i\gamma & 0\\
    0 & -i\gamma & \delta & 0\\
    -i\gamma & 0 & 0 & \delta
\end{pmatrix}$
\\
\hline
Pauli-Z &
\begin{tikzcd}
\meter{} 
\end{tikzcd}
&
$\begin{pmatrix}
    1 & 0 \\
    0 & -1 \\
\end{pmatrix}$
 \\
 \hline
NOT &
\begin{tikzcd}
\qw & \gate{X} & \qw
\end{tikzcd}
&
$\begin{pmatrix}
    0 & 1 \\
    1 & 0 \\
\end{pmatrix}$
 \\
 \hline
 \rule{0pt}{11ex}
Toffoli (CCNOT)
 &
 \begin{tikzcd}
    \qw & \ctrl{1} & \qw \\
    \qw & \octrl{1} & \qw \\
    \qw & \targ{} & \qw
 \end{tikzcd}
 &
 $\text{DIAG}(\bm{I}_4,X,\bm{I}_2)$
 \rule[-5ex]{0pt}{4ex}
 \\
 \hline
\end{tabular}
\end{center}
\end{table}

The GFL module leverages the graph structure of HSIs by propagating information at the superpixel level.
However, to preserve the detailed spatial information not retained in the superpixels, a complementary module is employed to learn features at the pixel level.
With the success of QUEEN in addressing the highly challenging HSI restoration problem \cite{lin2023hyperqueen}, the utility of quantum unitary-computing features generated by quantum neurons of QUEEN has been well demonstrated in recent hyperspectral remote sensing literature. 
Unlike the traditional deep learning that basically returns affine-computing features, the quantum unitary-computing features provide radically new information for better decision-making in the final detection stage.

Accordingly, we design a quantum feature learning module, QFL, to capture pixel-level information.
As shown in Fig. \ref{fig:QFL}, the QFL module is sequentially composed of a $1 \times 1$ convolutional layer, a QUEEN architecture, and a $1 \times 1$ convolutional layer.
The first $1 \times 1$ convolutional layer condenses the feature of the input differential feature map $\bX_\textnormal{diff}$ and reduces the feature map size, followed by the QUEEN to extract the quantum unitary-computing features.
Finally, a $1 \times 1$ convolutional layer is adopted to obtain the pixel-level feature map, which can be explicitly written as
\begin{equation*}
    f_\textnormal{QFL}(\bX_\textnormal{diff})=f_\textnormal{Conv}^{1\times 1}(f_\textnormal{QUEEN}(f_\textnormal{Conv}^{1\times 1}(\bX_\textnormal{diff}))),
\end{equation*}
where $f_\textnormal{QUEEN}$ is constructed in Fig. \ref{fig:QFL}.
Note that all the quantum neurons (all implemented via some Hamiltonian for a specific time) used to deploy the QUEEN must be some unitary operators, as the Schrödinger equation implies that the operators corresponding to those quantum neurons are all unitary \cite{Qunitary}.

To effectively extract useful pixel-level entangled quantum features using the quantum unitary operators, the QUEEN designed to have the mathematically provable FE, which means the quantum network can generate any possible quantum state or perform any unitary transform \cite{lin2023hyperqueen}, is adopted to explore the optimal features for addressing the HCD problem.
The quantum network is illustrated in Fig. \ref{fig:QFL} with the used quantum logic gates (neurons) defined in Table \ref{tab:common_qu_gate}, including rotation gates, the Ising gate, the Pauli gate, and the Toffoli gate.
As illustrated in Fig. \ref{fig:QFL}, the first layer of the QUEEN with FE encodes the classical feature map into quantum states using the angle embedding mechanism \cite{weigold2021encoding} for the subsequent quantum processing.
To be specific, the feature map $\bX_\textnormal{diff}$ is divided into groups of four elements along the channel direction (cf. $x_i$'s in Fig. \ref{fig:QFL}), and each group is independently fed into the quantum network as the angles of the $R_Y$ gate.
Next, a structure of ``$R_Y-XX-R_X-XX-R_Y$" is selected for the trainable quantum module to achieve mathematically provable full expressibility (FE), where FE can be guaranteed by the following theorem.
\begin{Theorem} \label{theorem: FE}
There exist some real-valued trainable network parameters $\{\rho_{k},\omega_{k},\theta_{k},\phi_{k}\}$, such that the trainable quantum neurons deployed in the proposed QUEEN (cf. Fig. \ref{fig:QFL}) can express all valid quantum unitary operators.
\hfill$\square$
\end{Theorem}
The proof of this FE theorem directly follows from Theorem 2 of \cite{lin2023hyperqueen}, and is hence omitted here.
After the core FE quantum processing stage, Toffoli gates are used as the entanglement gates for capturing the quantum entanglement effect.
Finally, the quantum Pauli measurements performed only on the odd qubits mimic the max pooling mechanism in the traditional deep learning, and accordingly obtain the pixel-level quantum unitary-computing feature map for the subsequent detection.

\subsection{QEC and Loss Function}\label{sec: Qclass}

\begin{figure}[t]
    \centering
    \includegraphics[width=1\linewidth]{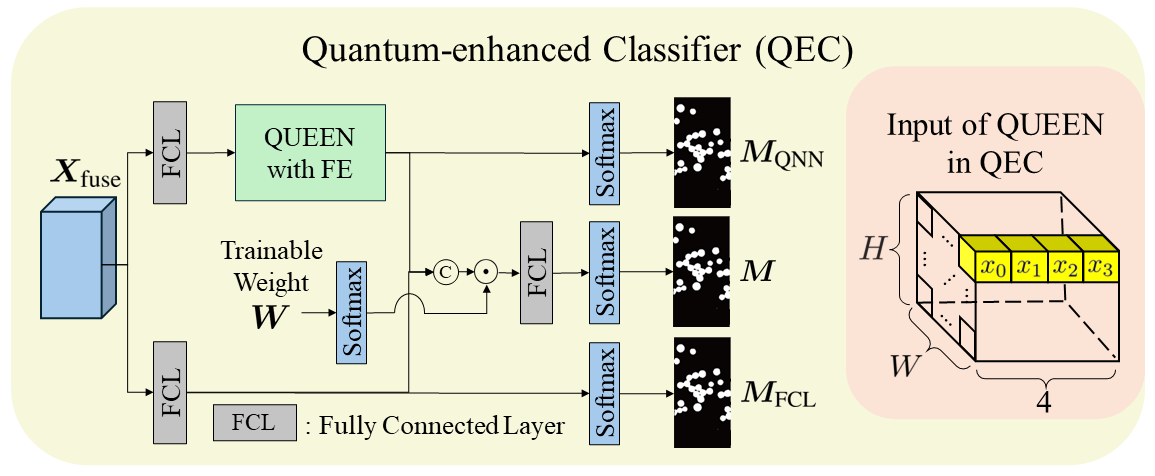}
    \caption{Detailed architecture of the QEC module with the input dimension of the QUEEN.
    QEC first learns the information separately through a quantum network and a traditional fully connected layer.
    Then, a trainable weight $\bW$ highlights the important elements, followed by another fully connected layer to fuse the quantum and the traditional features.
    Finally, the output feature map $\bM$ is computed by the Softmax function.
    The block at the right illustrates how to input the feature map into the QUEEN to learn the quantum information.
    }\label{fig:QEC}
\end{figure}

Motivated by the outstanding performances of QNNs in recent literature \cite{lin2023hyperqueen,fan2023hybrid, senokosov2024quantum}, we introduce a quantum-enhanced classifier, QEC, which fuses QNN and the traditional classifier, for the final classification stage.
QEC fuses the affine-mapping information generated by the traditional classifier with the entangled unitary-computing information learned from QUEEN, integrating fundamentally different perspectives to achieve a better detection performance.
As shown in Fig. \ref{fig:QEC}, the QEC first divides the process into two distinct branches (i.e., a quantum branch and a traditional branch), each utilizing a fundamentally different mechanism to detect changed pixels, followed by sharing information with each other through a fusion procedure.
As illustrated in Fig. \ref{fig:QEC}, the quantum classifier consists of a fully connected layer and a QUEEN with FE, whose structure is identical to the QUEEN used in the QFL module.
In contrast, the traditional classifier consists solely of a fully connected layer.
After feeding the fused feature map $\bX_\textnormal{fuse}$ generated from the previous layer, the outputs of the two branches are concatenated, followed by elementwise multiplying a trainable attention weight $\bW \in\mathbb{R}^{H\times W\times 4}$, which is normalized by the Softmax function, to highlight those important elements.
Finally, the quantum feature map and the traditional feature map are fused by a fully connected layer, followed by the Softmax function to generate the output change detection map $\bM$.
The process of the QEC module to produce $\bM$ can be progressively expressed as
\begin{align*}
   \bX_\textnormal{QNN}=&f_\textnormal{QUEEN}(f_\textnormal{FCL}^{(1)}(\bX_\textnormal{fuse})),\\
   \bX_\textnormal{FCL}=&f_\textnormal{FCL}^{(2)}(\bX_\textnormal{fuse}),\\
   \bM=&f_\textnormal{Softmax}(f_\textnormal{FCL}^{(3)}((\bX_\textnormal{QNN}\|\bX_\textnormal{FCL})\odot f_\textnormal{Softmax}(\bW))),
\end{align*}
and the three $f_\textnormal{FCL}$ modules do not share weights as shown in Fig. \ref{fig:QEC}.
By progressively following the aforementioned process, the QEC can be elegantly implemented as shown in Fig. \ref{fig:QEC}.

Given the ground-truth change map $\bM_{\textnormal{GT}}$, cross-entropy (CE) is selected as the loss function, and its definition is
\begin{equation}
 L_\textnormal{CE}(\bM_{\textnormal{GT}},\bM):=-\frac{1}{N} \sum_{i=1}^N \left( [{\bm m}_{\textnormal{GT}}]_{i} \log \left([{\bm m}]_i\right)\right), \label{eq:CEloss}
\end{equation}
where $N$ is the total number of training samples, and $[{\bm m}_{\textnormal{GT}}]_{i}$ and 
$[{\bm m}]_i$ are the $i$th training samples from $\bM_{\textnormal{GT}}$ and $\bM$, respectively.
Additionally, we aim to enhance the discriminative capability of the outputs generated by the QNN branch and the traditional branch to improve performance.
Therefore, besides the CE loss between $\bM_{\textnormal{GT}}$ and $\bM$, an auxiliary term $L_\textnormal{aux}$ based on the CE loss defined in \eqref{eq:CEloss} for ensuring the discriminative capability of $\bX_\textnormal{QNN}$ and $\bX_\textnormal{FCL}$ is incorporated into the loss function.
Thus, the final loss function $L$ can then be explicitly expressed as
\begin{align*}
   L_\textnormal{aux}&=L_\textnormal{CE}(\bM_{\textnormal{GT}},\bM_{\textnormal{QNN}})  +L_\textnormal{CE}(\bM_{\textnormal{GT}},\bM_{\textnormal{FCL}}),
   \\
   L&= L_\textnormal{CE}(\bM_{\textnormal{GT}},\bM)+\frac{1}{2} L_\textnormal{aux},
\end{align*}
where $\bM_\textnormal{QNN}\triangleq f_\textnormal{Softmax}(\bX_\textnormal{QNN})$ and $\bM_\textnormal{FCL}\triangleq f_\textnormal{Softmax}(\bX_\textnormal{FCL})$, as illustrated in Fig. \ref{fig:QEC}.
With the auxiliary term to ensure the discriminative capability of both the quantum and traditional classifiers, we can further upgrade the performance of the proposed QUEEN-$\mathcal{G}$.

To summarize, we have proposed a QUEEN-empowered graph neural network, termed as QUEEN-$\mathcal{G}$, for the HCD problem.
At the feature extraction stage, GFL and QFL extract fundamentally different features.
GFL learns affine-mapping information at the superpixel level.
In parallel, QFL works under the pixel-level unitary-computing mechanism for preserving the detailed spatial information not retained in the superpixels, serving as a complement to the superpixel-level GFL module.
Moreover, at the final classification stage, QEC is proposed to fuse the quantum classifier and the traditional fully connected layer.
For both the feature extraction stage and the classification stage, two types of information are learned from two radically different perspectives, and this strategy helps the proposed QUEEN-$\mathcal{G}$ to enhance the final detection performance through the gained novel information.
The effectiveness of the quantum modules QFL and QEC in QUEEN-$\mathcal{G}$ will be illustrated in Section \ref{sec:experiment}.

\section{Experimental Results}\label{sec:experiment}

In this section, we demonstrate the superiority of the proposed QUEEN-$\mathcal{G}$ on real hyperspectral datasets by comparing it with state-of-the-art HCD algorithms.
Specifically, Section \ref{sec:dataset} provides an overview of the three real hyperspectral datasets used in the experiment, along with the definitions of the five quantitative metrics employed for evaluation.
The benchmark HCD methods and the experimental setup are detailed in Section \ref{sec:setting}.
In Section \ref{sec:performance}, we present a detailed analysis of the performance of the proposed QUEEN-$\mathcal{G}$ compared to the benchmark HCD algorithms.
In Section \ref{sec:discussion}, the analyses of the proposed algorithm are discussed, including an ablation study of the effectiveness of the proposed QFL and QEC in Section \ref{sec:ablation}, an experiment assessing the impact of varying sampling rates on the performance of QUEEN-$\mathcal{G}$ in Section \ref{sec:rate}, an experiment investigating the parameter sensitivity of the proposed algorithm in Section \ref{sec:parameter}, and an analysis of the computational efficiency of QUEEN-$\mathcal{G}$ in Section \ref{sec:computational}.

\subsection{Hyperspectral Datasets and Quantitative Assessments}\label{sec:dataset}

\begin{figure}[t]
\centering
    \subfigure[]{
        \includegraphics[width=.29\linewidth]{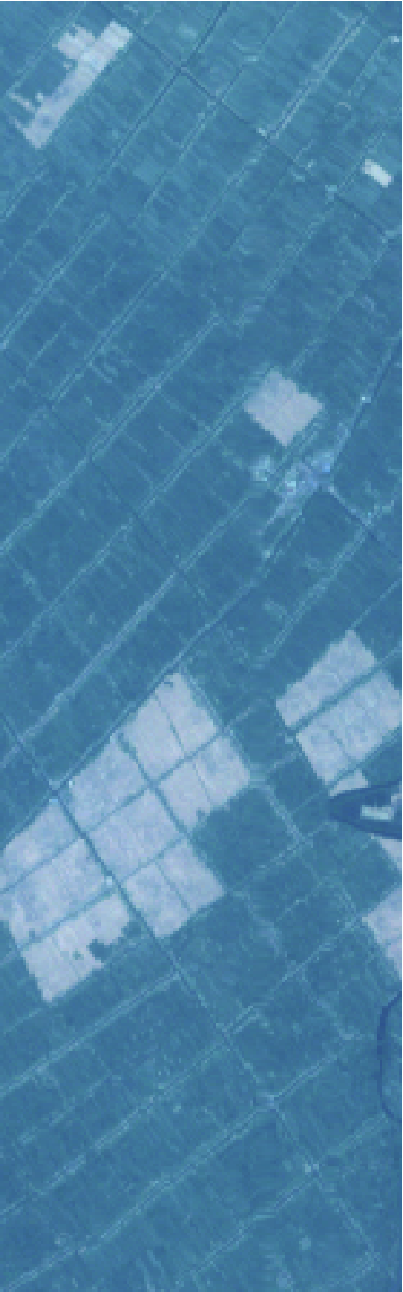}
        \label{fig:Yan_t1}
    }
    \subfigure[]{
        \includegraphics[width=.29\linewidth]{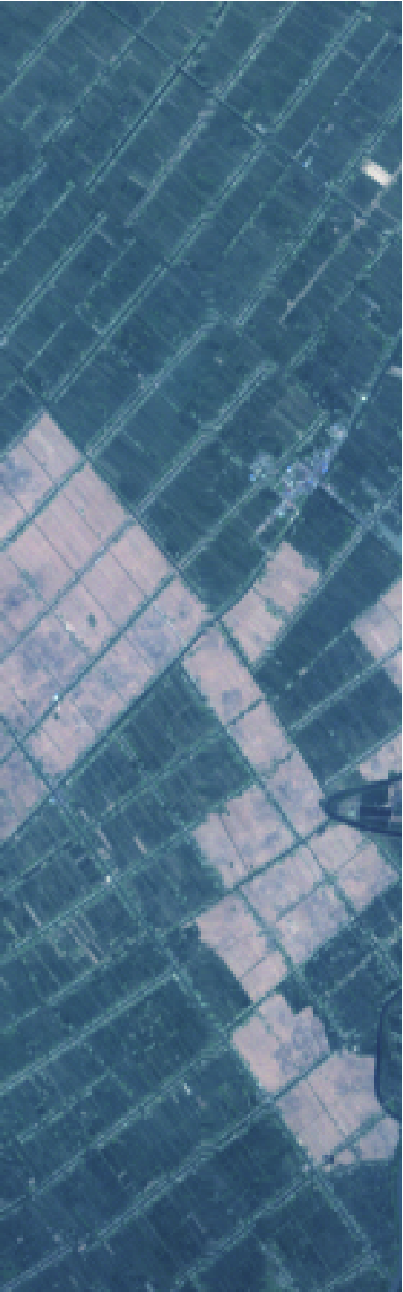}
        \label{fig:Yan_t2}
    }
    \subfigure[]{
        \includegraphics[width=.29\linewidth]{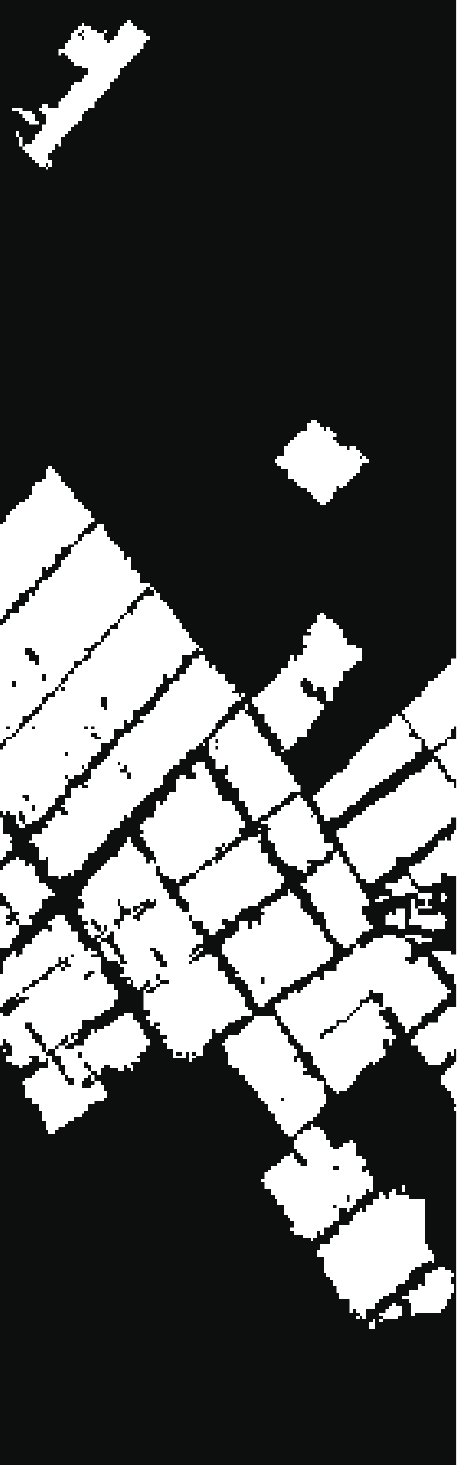}
        \label{fig:Yan_gt}
    }
    \caption{Pseudo-color images of the benchmark Yancheng dataset. (a) HSI acquired on May 3rd, 2006. (b) HSI acquired on April 23rd, 2007. (c) Ground-truth change map, where white pixels denote the changes.}\label{fig:Yan}
\end{figure}

\begin{figure}[t]
\centering
    \subfigure[]{
        \includegraphics[width=.29\linewidth]{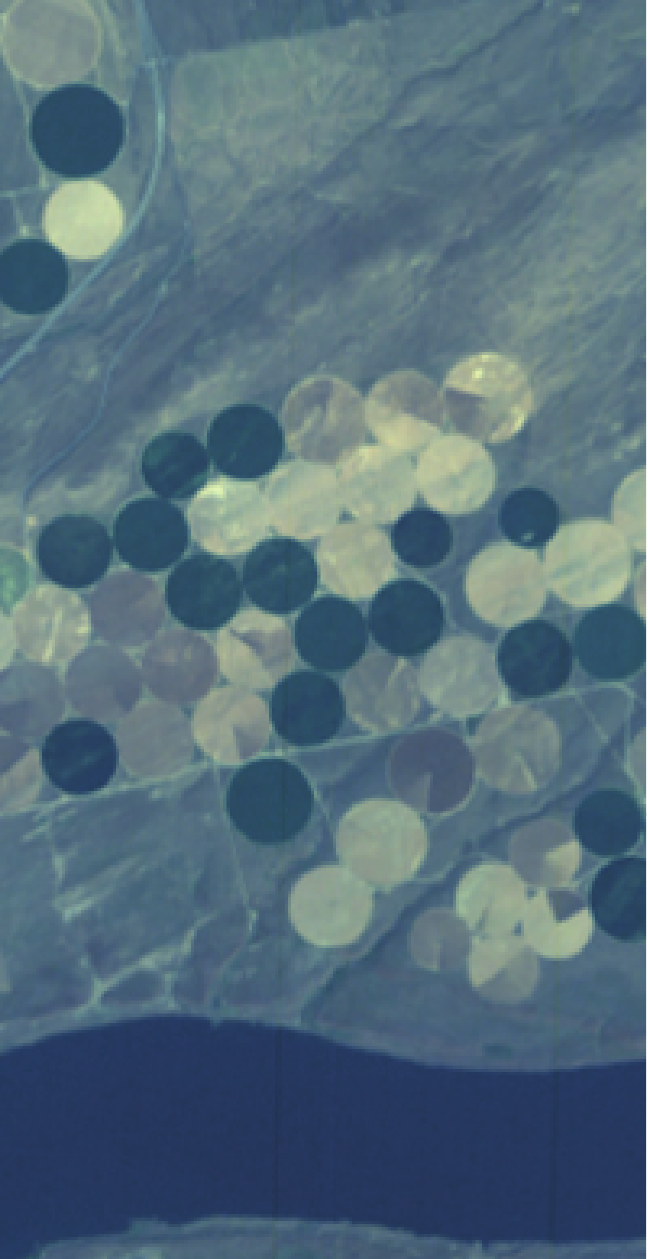}
        \label{fig:Her_t1}
    }
    \subfigure[]{
        \includegraphics[width=.29\linewidth]{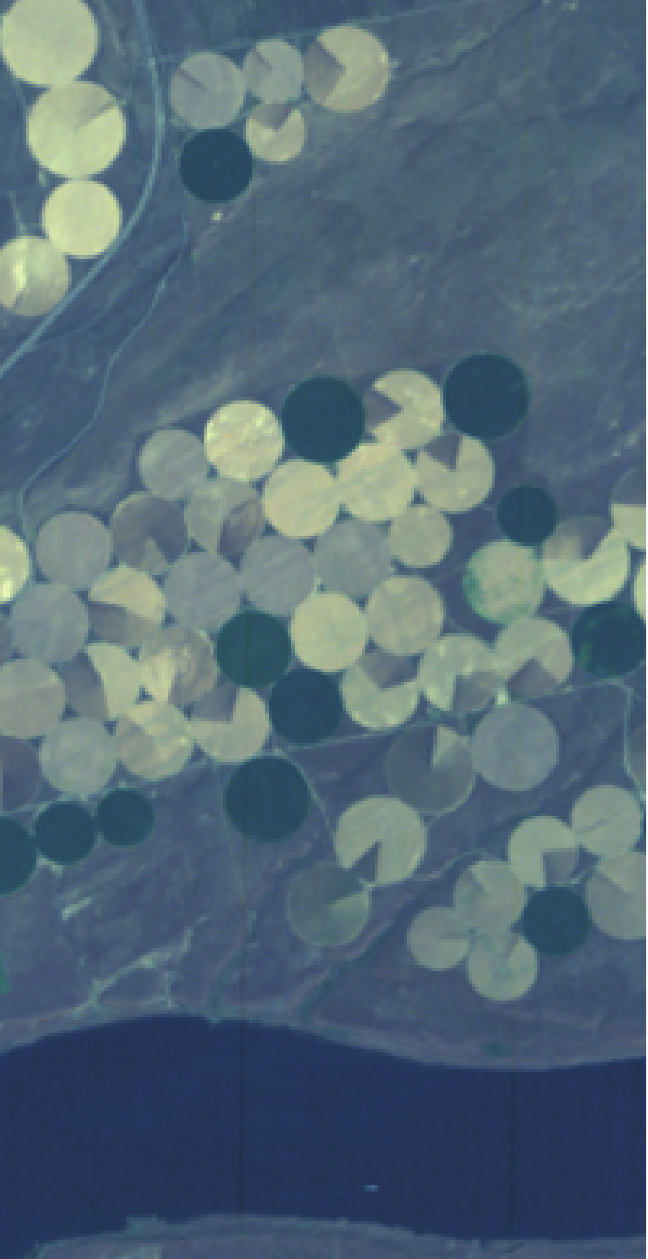}
        \label{fig:Her_t2}
    }
    \subfigure[]{
        \includegraphics[width=.29\linewidth]{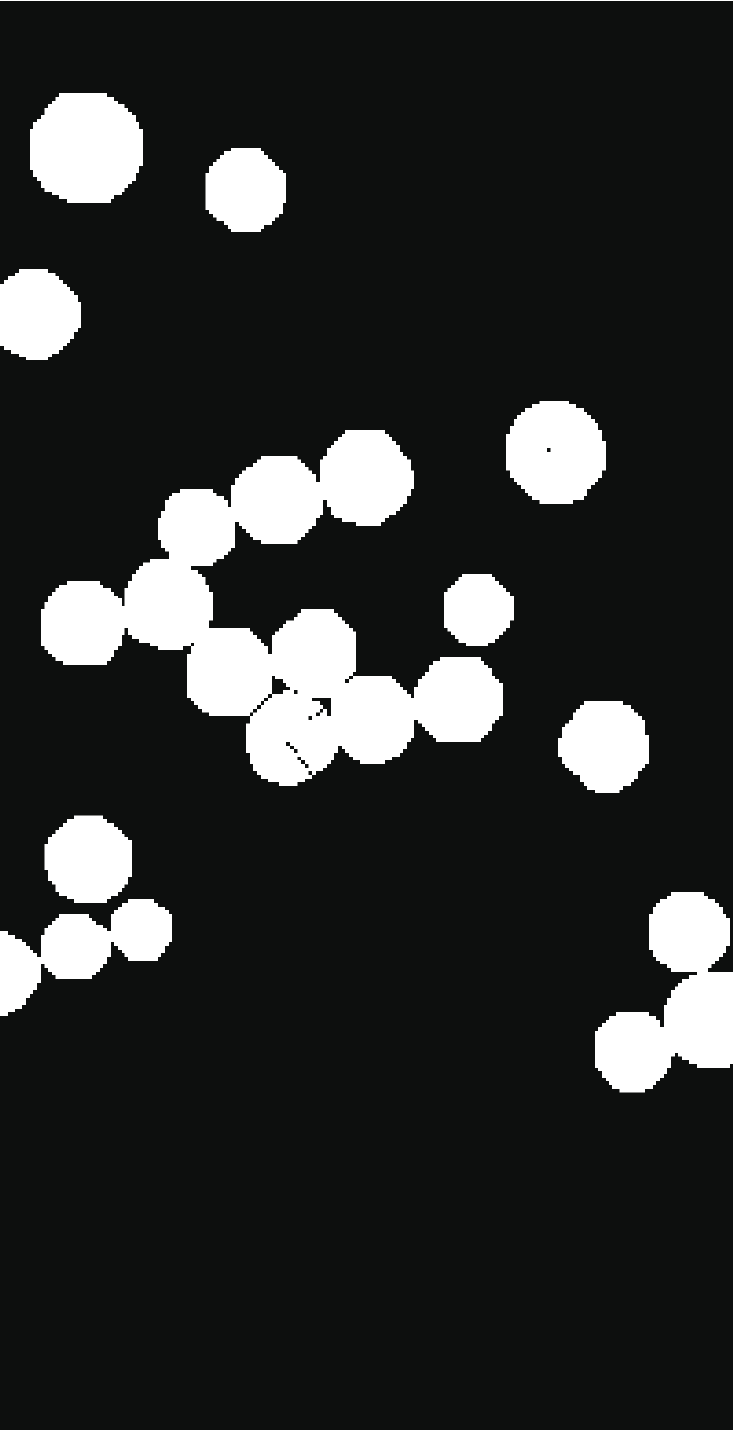}
        \label{fig:Her_gt}
    }
    \caption{Pseudo-color images of the benchmark Hermiston dataset. (a) HSI acquired on May 1st, 2004. (b) HSI acquired on May 8th, 2007. (c) Ground-truth change map, where white pixels denote the changes.}
    \label{fig:Her}
\end{figure}
\begin{figure}[t]
\centering
    \subfigure[]{
        \includegraphics[width=.29\linewidth]{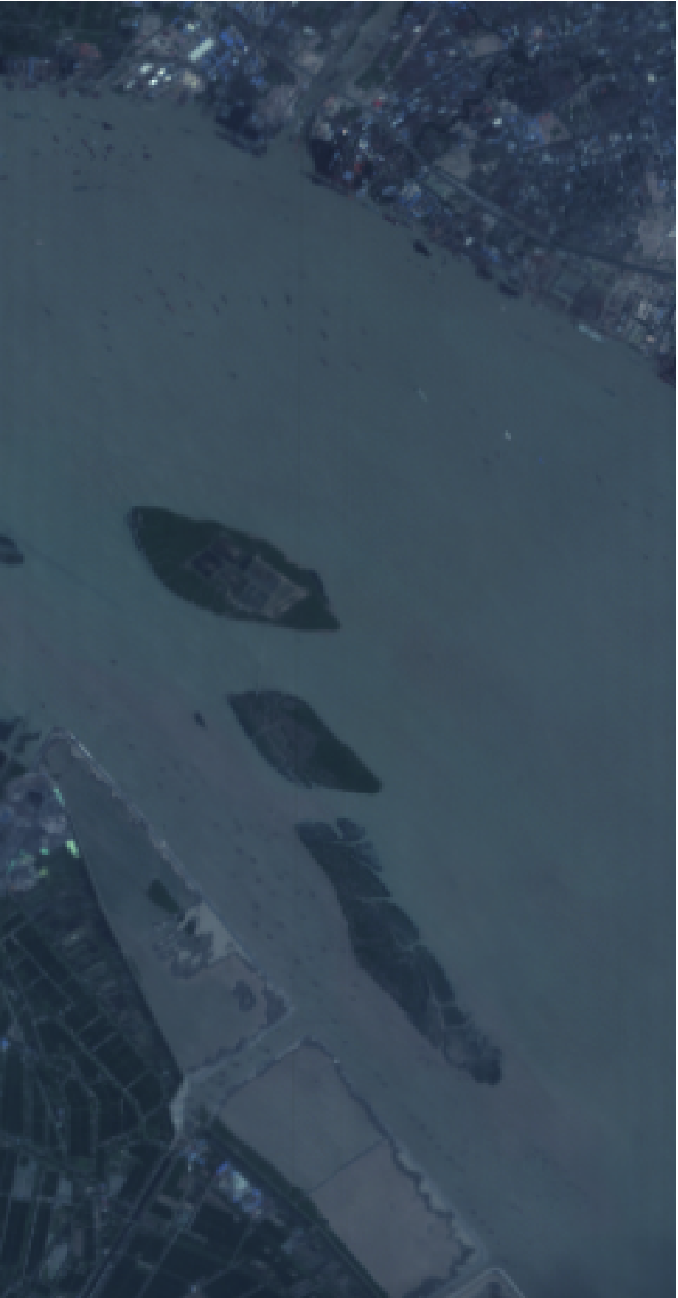}
        \label{fig:Jia_t1}
    }
    \subfigure[]{
        \includegraphics[width=.29\linewidth]{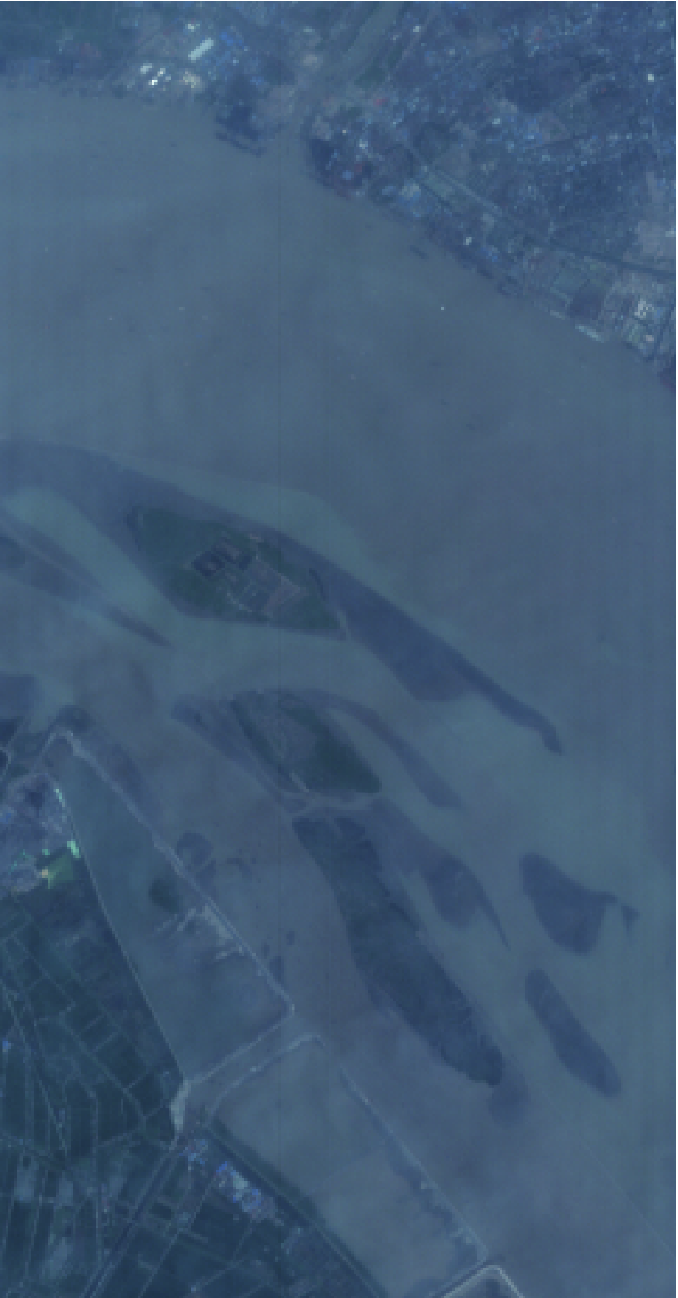}
        \label{fig:Jia_t2}
    }
    \subfigure[]{
        \includegraphics[width=.29\linewidth]{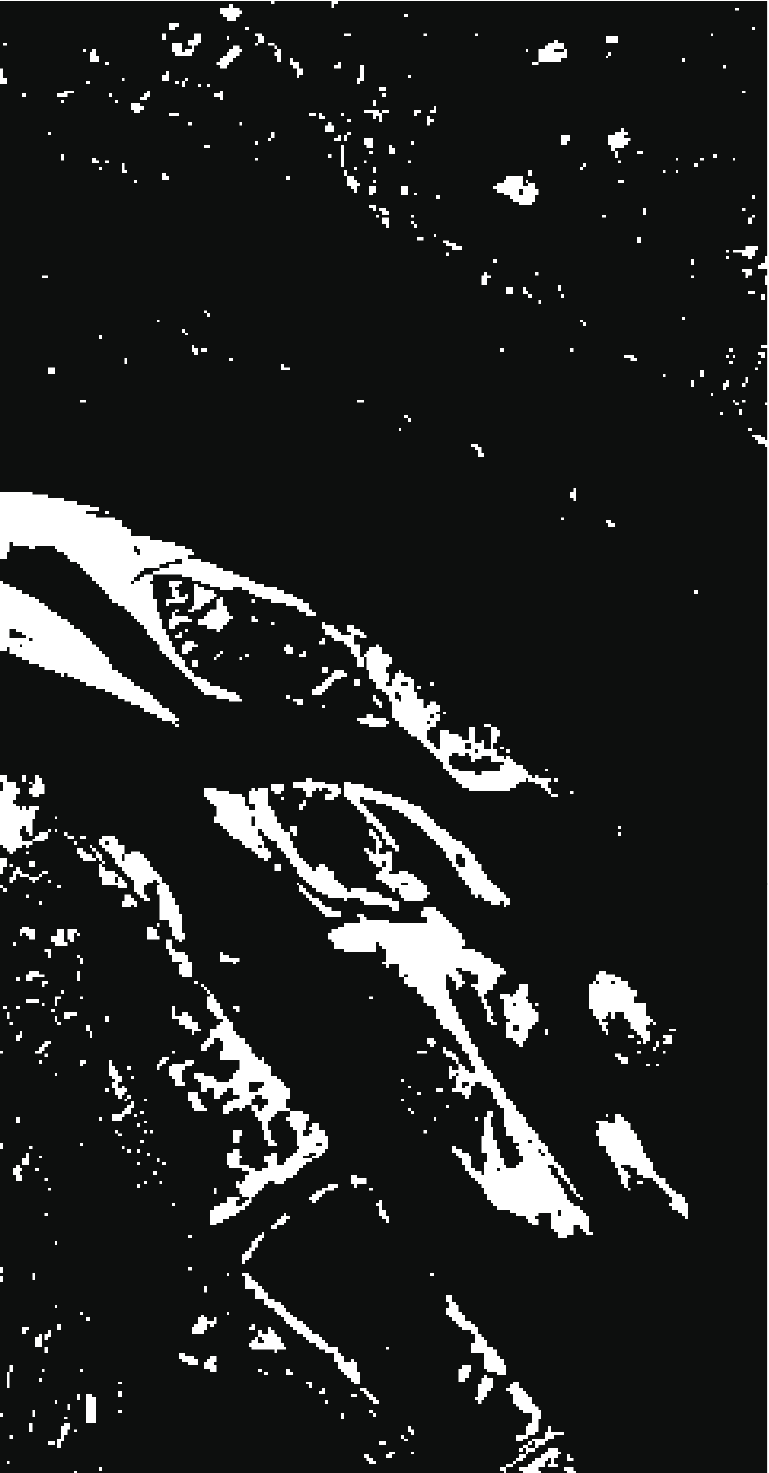}
        \label{fig:Jia_gt}
    }
    \caption{Pseudo-color images of the benchmark Jiangsu dataset. (a) HSI acquired on May 3rd, 2013. (b) HSI acquired on December 31st, 2013. (c) Ground-truth change map, where white pixels denote the changes.}\label{fig:Jia}
\end{figure}

In this study, three extensively utilized hyperspectral datasets are adopted to compare the proposed QUEEN-$\mathcal{G}$ with other state-of-the-art HCD methods.
The evaluation is conducted using five representative quantitative indices.
Each dataset consists of a pair of bitemporal HSIs and a corresponding ground-truth image.
The bitemporal HSIs encompass the same geographical area but were acquired on different dates.
At the same time, the ground-truth images provide detailed information regarding the spatial distribution of both changed and unchanged pixels within the observed region.

    \subsubsection{Yancheng Dataset} 
    The Yancheng dataset is captured by Hyperion sensor over a wetland agricultural area in Yancheng, Jiangsu Province, China.
    The bitemporal images were taken on May 3, 2006, and April 23, 2007, encompassing 242 spectral bands from 350 to 2580 nm \cite{liu2017band}.
    In this paper, bitemporal HSIs containing spatial dimensions  $H \times W=450\times140$ pixels and $C=155$ bands after noise removal are utilized for the experiments.
    The dataset primarily highlights agricultural changes involving vegetation, bare land, water, and soil \cite{liu2019unsupervised}.
    Fig. \ref{fig:Yan} presents pseudo-color representations of the bitemporal Yancheng dataset, accompanied by the corresponding ground-truth change map.

    \subsubsection{Hermiston Dataset}
    The Hermiston dataset is captured by Hyperion sensor over Hermiston city in Umatilla County, United States.
    The bitemporal images were taken on May 1, 2004, and May 8, 2007,
    encompassing 242 spectral bands from 350 to 2580 nm\cite{guo2021change}.
    In this paper, bitemporal HSIs containing spatial dimensions $H \times W=390\times 200$ pixels and $C=154$ bands after noise removal are utilized for the experiments.
    This dataset primarily showcases agricultural changes, including shifts between crops, soil, water, and other land cover types.
    Fig. \ref{fig:Her} presents pseudo-color representations of the bitemporal Hermiston dataset, accompanied by the corresponding ground-truth change map.

    \subsubsection{Jiangsu Dataset}
    The Jiangsu dataset is captured by Hyperion sensor over Jiangsu Province, China.
    The bitemporal images were taken on May 3 and December 31, 2013, encompassing 242 spectral bands from 350 to 2580 nm\cite{wang2018getnet}.
    In this paper, bitemporal HSIs containing spatial dimensions $H \times W=463\times241$ pixels and $C=128$ bands after noise removal are utilized for the experiments.
    The primary change detected in this dataset is the disappearance of substances within the river.
    Fig. \ref{fig:Jia} presents pseudo-color representations of the bitemporal Jiangsu dataset, accompanied by the corresponding ground-truth change map.

To conduct a fair comparison of the HCD methods, five representative quantitative metrics are employed, i.e., overall accuracy (OA), kappa coefficient $\kappa$, F1-score (F1), precision (Pr), and recall (Re).
Before discussing these quantitative indices in detail, it is important to define the fundamental terms TP, FP, TN, and FN, standing for true positives, false positives, true negatives, and false negatives, respectively.
Overall accuracy (OA) represents the proportion of correctly detected pixels and is mathematically expressed as
\begin{equation*} 
\text{OA}=\frac{\text{TP}+\text{TN}}{\text{TP}+\text{TN}+\text{FP}+\text{FN}}.
\end{equation*}
The kappa coefficient $\kappa$ is a statistical measure that accounts for the inter-rater reliability, offering a more reliable assessment than OA.
To compute $\kappa$, we first define the hypothetical probability of chance agreement $\text{P}_e$ as
 \begin{equation*} 
\text{P}_e=\frac{(\text{TP} + \text{FP})(\text{TP} + \text{FN}) + (\text{FN} + \text{TN})(\text{FP} + \text{TN)}}{(\text{TP} + \text{TN} + \text{FP} + \text{FN})^{2}}.
\end{equation*}
Subsequently, $\kappa$ is determined as
\begin{equation*} 
\kappa=\frac{\text{OA}-\text{P}_e}{1-\text{P}_e}.
\end{equation*}
F1-score is a metric that balances Pr and Re, particularly useful in scenarios involving imbalanced data.
Precision (Pr) measures the proportion of correctly identified positive instances among all predicted positives, defined as
\begin{equation*} 
\text{Pr}=\frac{\text{TP}}{\text{TP}+\text{FP}}.
\end{equation*}
Recall (Re), on the other hand, assesses the proportion of actual positives correctly identified by the algorithm and is defined as
\begin{equation*} 
\text{Re}=\frac{\text{TP}}{\text{TP}+\text{FN}}.
\end{equation*}
Finally, the F1-score (F1) is defined as $\text{F1}=2 \times \frac{\text{Pr}\times\text{Re}}{\text{Pr}+\text{Re}}$, which is the harmonic mean of precision and recall.
%
\begingroup
\setlength{\tabcolsep}{6pt} 
\renewcommand{\arraystretch}{1.2} 
\begin{table*}[t]
\begin{center}
\caption{Quantitative comparison on the three real hyperspectral datasets.}\label{tab1:result}
	\scalebox{1.15}{
\begin{tabular}{cc|ccccccc}
\hline
\hline
Dataset & Index & ML-EDAN & SST-Former & MSDFFN & HyGSTAN & $\textnormal{D}^2$AGCN & CSDBF  & QUEEN-$\mathcal{G}$\\
\hline
\multirow{5}{*}{Yancheng} & OA($\uparrow$) & 0.962 &  {\bf 0.967} & 0.961 & 0.961 & 0.923 & 0.961 
 & {\bf \underline{0.969}}\\
& $\kappa$($\uparrow$) & 0.909 & {\bf 0.920} & 0.909 & 0.908 & 0.823 & 0.907 
 & {\bf \underline{0.925}}\\
 & F1($\uparrow$) & 0.936 & {\bf 0.944} & 0.936 & 0.935 & 0.878 & 0.934 
 & {\bf \underline{0.947}}\\
& Pr($\uparrow$) & 0.905 & {\bf 0.915} & 0.897 & 0.902 & 0.801 & 0.905 
 & {\bf \underline{0.926}}\\
& Re($\uparrow$) & 0.970 & {\bf 0.975} & {\bf \underline{0.979}} & 0.971 & 0.973 & 0.966 
 & 0.970\\
\hline
\multirow{5}{*}{Hermiston} & OA($\uparrow$) & 0.980  & 0.982 & \textbf{0.983} & 0.980 & 0.962 & 0.983 
 & {\bf \underline{0.986}}\\
& $\kappa$($\uparrow$) & 0.917  & 0.921 & \textbf{0.927} & 0.914 & 0.844 & \textbf{0.927} & {\bf \underline{0.937}}\\
& F1($\uparrow$) & 0.928 & 0.932 & \textbf{0.937} & 0.926 & 0.866 & 0.936 & {\bf \underline{0.945}}\\
& Pr($\uparrow$) & 0.875 & 0.885 & 0.887 & 0.871 & 0.787 & {\bf 0.889} & {\bf \underline{0.905}}\\
& Re($\uparrow$) & 0.989 & 0.983 & {\bf \underline{0.993}} & 0.989 & 0.966 & 0.988 & \textbf{0.990}\\
\hline
\multirow{5}{*}{Jiangsu} & OA($\uparrow$) & 0.910  & 0.934 & 0.919 & 0.883 & 0.876 & \textbf{0.938} & {\bf \underline{0.943}}\\
& $\kappa$($\uparrow$) & 0.598 & 0.673 & 0.632 & 0.508 & 0.484 & \textbf{0.685} & {\bf \underline{0.706}}\\
& F1($\uparrow$) & 0.643 & 0.707 & 0.674 & 0.564 & 0.543 & \textbf{0.718} & {\bf \underline{0.736}}\\
& Pr($\uparrow$) & 0.492 & 0.562 & 0.520 & 0.412 & 0.391 & \textbf{0.577} & {\bf \underline{0.608}}\\
& Re($\uparrow$) & 0.929 & \textbf{0.953} & {\bf \underline{0.960}} & 0.903 & 0.888 & 0.949 & 0.941\\
\hline
\hline

\end{tabular}}
\end{center}
\end{table*}  
\endgroup

\subsection{Peer Methods and Experimental Setup}\label{sec:setting}

In this paper, we compare the proposed QUEEN-$\mathcal{G}$ with six benchmark HCD algorithms, including ML-EDAN \cite{qu2021multilevel}, SST-Former \cite{wang2022spectral}, MSDFFN \cite{luo2023multiscale}, HyGSTAN \cite{yu2024hyperspectral}, $\textnormal{D}^2$AGCN \cite{qu2021dual}, and CSDBF \cite{wang2022csdbf}.
ML-EDAN is designed based on the encoder-decoder structure and the LSTM network.
SST-Former and MSDFFN are Transformer-based algorithms.
HyGSTAN is a lightweight architectural network that utilizes the attention mechanism.
$\textnormal{D}^2$AGCN and CSDBF are representative GNN-based methods.

All the experiments in this paper are conducted on the computer facility equipped with NVIDIA RTX-3090 GPU and AMD Ryzen 9 5950X CPU with 3.40-GHz speed and 128-GB random access memory.
The proposed network is implemented with Python 3.9.18, Pytorch 1.10.1, and PennyLane 0.37.0.
During the training phase, the Adam optimizer \cite{kingma2014adam} is adopted to optimize the proposed network.
The learning rate was initialized at 0.005 and decayed by a factor of 0.9 at every 20 epochs, and the maximum number of epochs was set at 250.
For all three datasets, a unified parameter setting of the superpixel segmentation scale is set to $s=20$, which consequently determines the number of superpixels as $HW/s$.
%
%
The training data is sampled at a rate of $1\%$ from all the pixels with a balanced ratio of 1:1 between changed and unchanged samples. 
This subset is subsequently partitioned into training and validation sets by a 9:1 ratio.
Considering the random nature of training deep networks, the results of all HCD methods in this paper represent the mean values obtained from 10 independent Monte Carlo runs.

\subsection{Quantitative and Qualitative Analyses}\label{sec:performance}

\begin{figure*}[t]
    \centering
    \subfigure[]{
        \includegraphics[width=0.106\linewidth]{final_fig/Yan_GT.pdf}
        \label{fig:Yan_GT}
    }
    \subfigure[]{
        \includegraphics[width=0.106\linewidth]{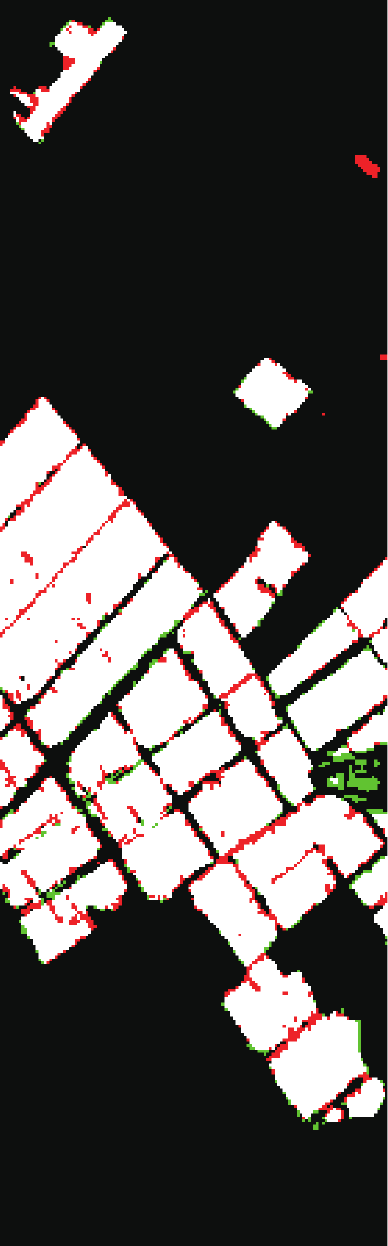}
        \label{fig:Yan_MLEDAN}
    }
    \subfigure[]{
        \includegraphics[width=0.106\linewidth]{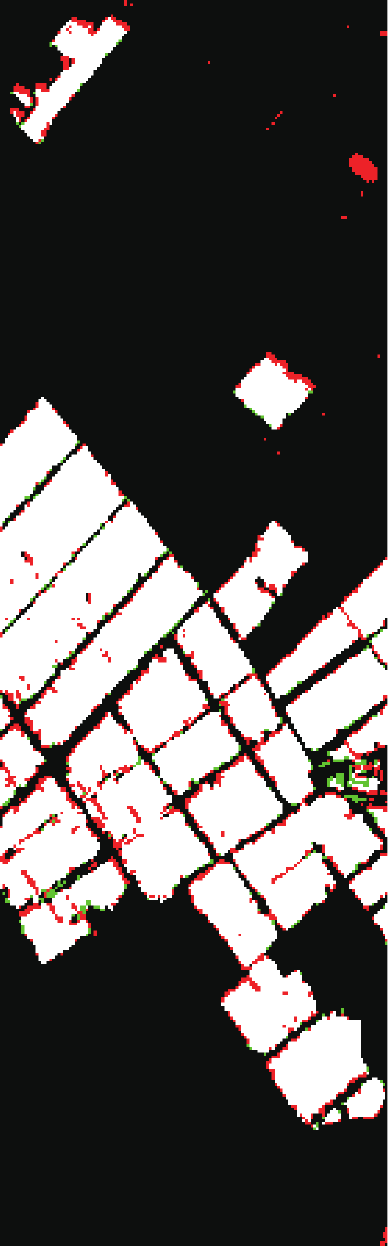}
        \label{fig:Yan_SSTF}
    }
    \subfigure[]{
        \includegraphics[width=0.106\linewidth]{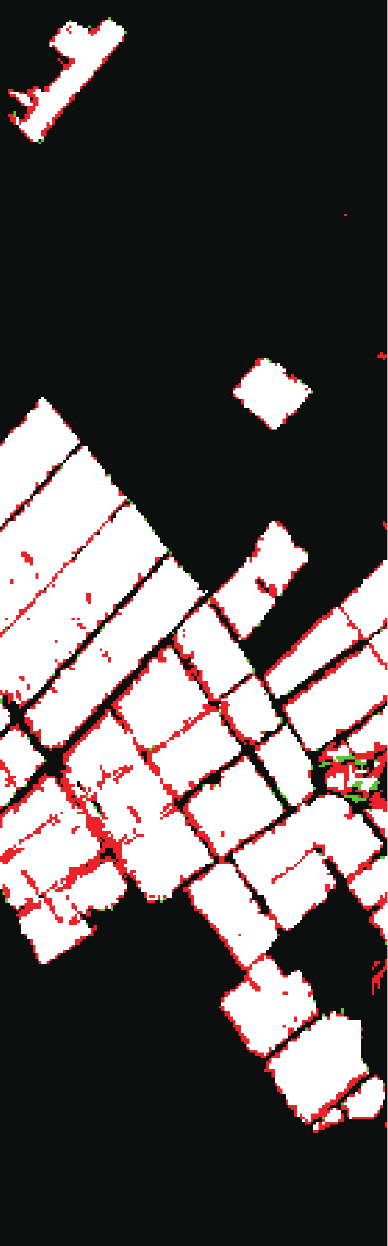}
        \label{fig:Yan_MSDFFN}
    }
    \subfigure[]{
        \includegraphics[width=0.106\linewidth]{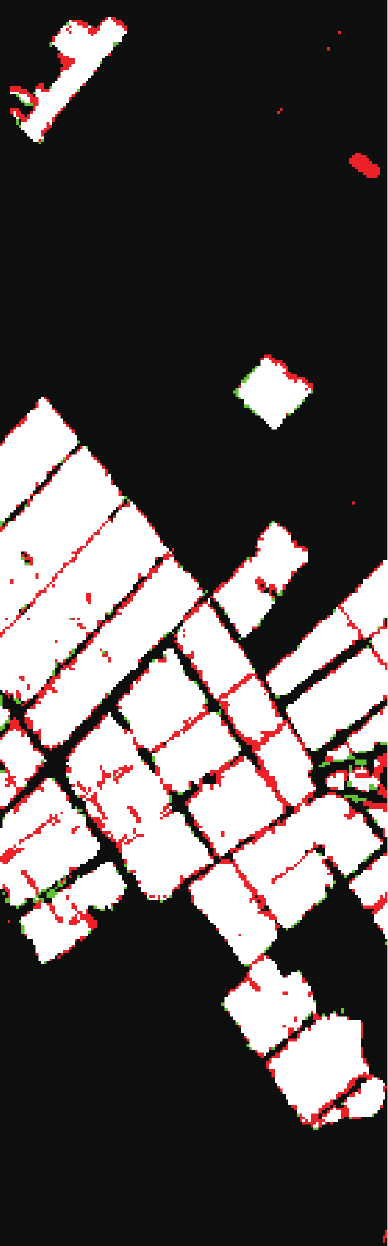}
        \label{fig:Yan_HyGSTAN}
    }
    \subfigure[]{
        \includegraphics[width=0.106\linewidth]{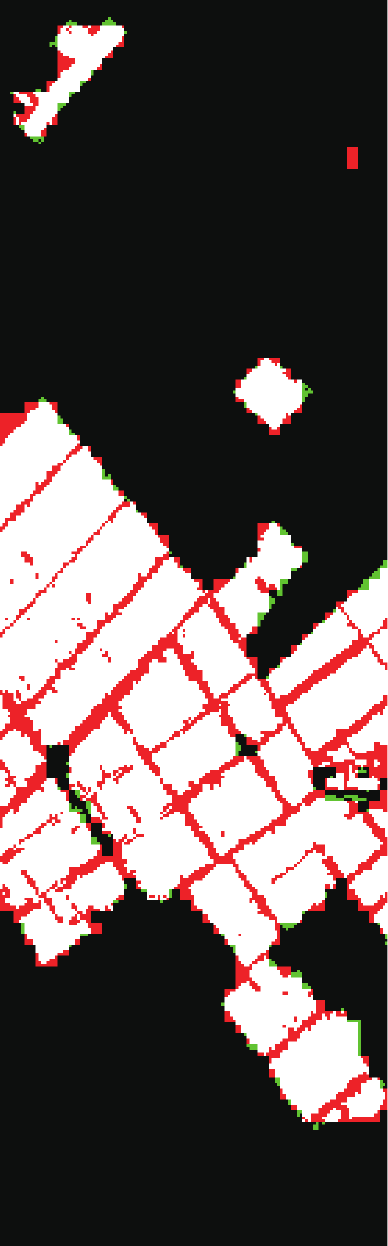}
        \label{fig:Yan_G2AGCN}
    }
    \subfigure[]{
        \includegraphics[width=0.106\linewidth]{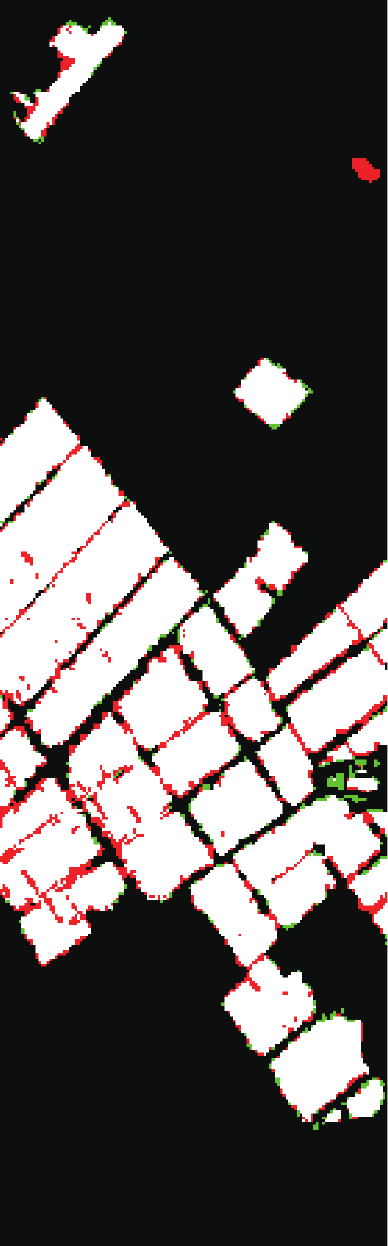}
        \label{fig:Yan_CSDBF}
    }
    \subfigure[]{
        \includegraphics[width=0.106\linewidth]{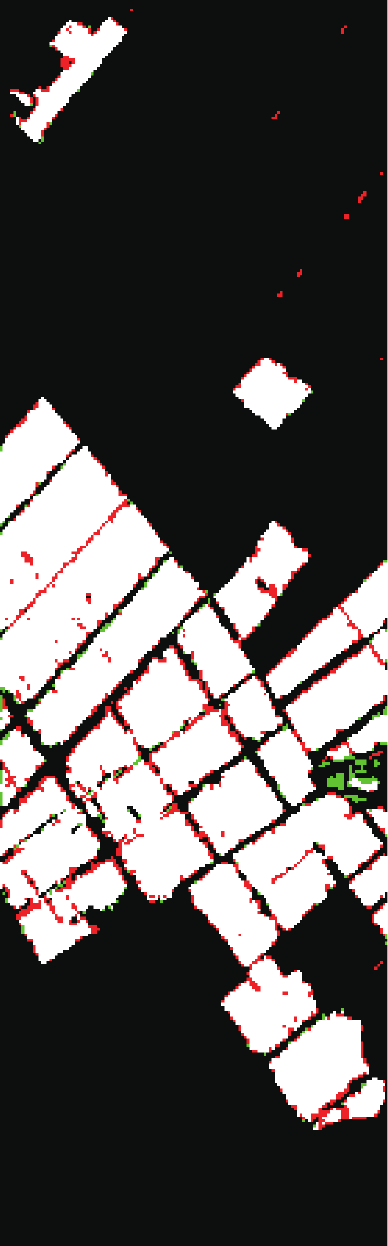}
        \label{fig:Yan_proposed}
    }
    \caption{Change detection results of the HCD algorithms on the Yancheng dataset with FP and FN marked in red and green, respectively. (a) Ground Truth. (b) ML-EDAN. (c) SST-Former. (d) MSDFFN. (e) HyGSTAN. (f) $\textnormal{D}^2$AGCN. (g) CSDBF. (h) QUEEN-$\mathcal{G}$.}
    \label{fig:Yan-exp}
    \vspace{-0.3cm}
\end{figure*}

\begin{figure*}[t]
    \centering
    \subfigure[]{
        \includegraphics[width=0.106\linewidth]{final_fig/Her_GT.pdf}
        \label{fig:Her_GT}
    }
    \subfigure[]{
        \includegraphics[width=0.106\linewidth]{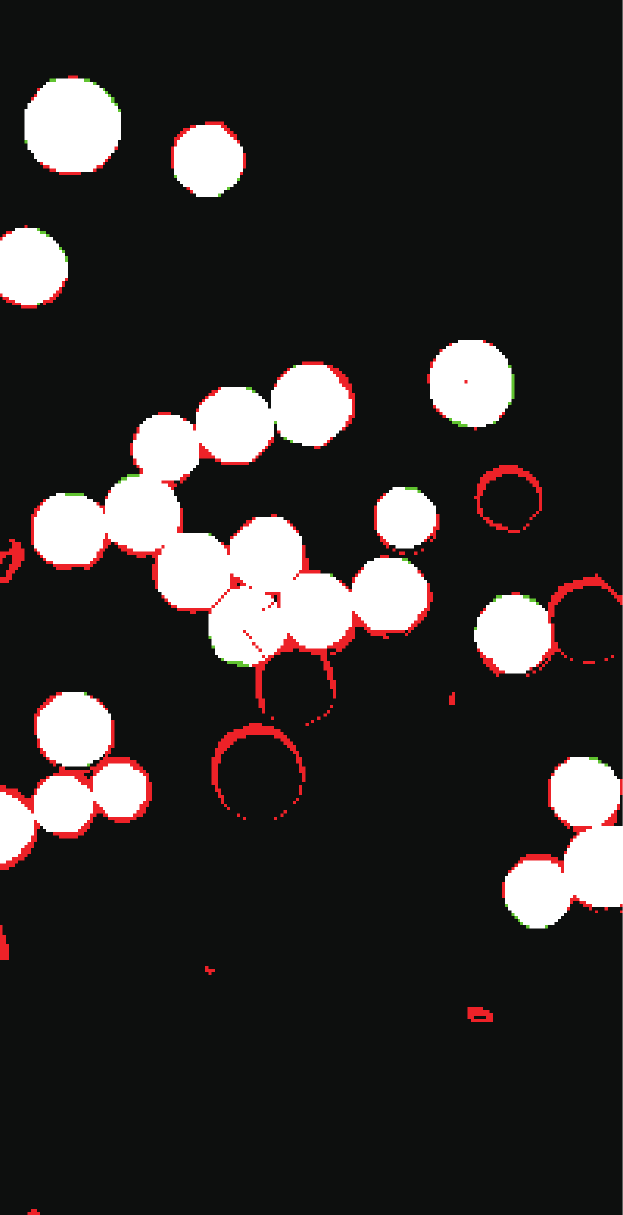}
        \label{fig:Her_MLEDAN}
    }
    \subfigure[]{
        \includegraphics[width=0.106\linewidth]{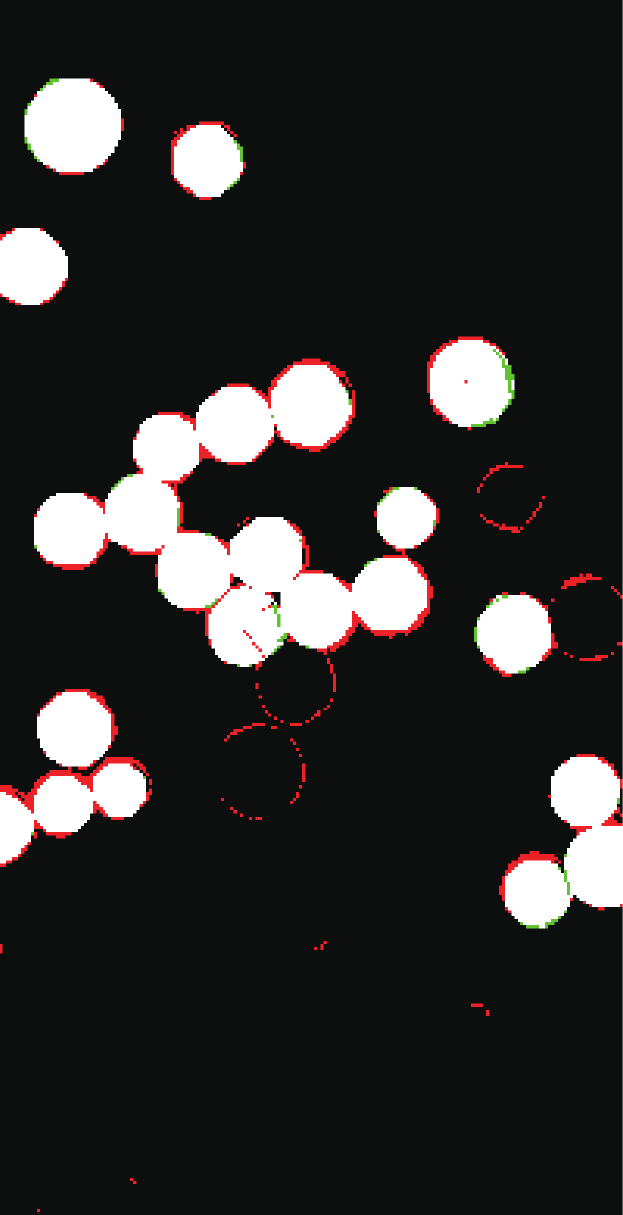}
        \label{fig:Her_SSTF}
    }
    \subfigure[]{
        \includegraphics[width=0.106\linewidth]{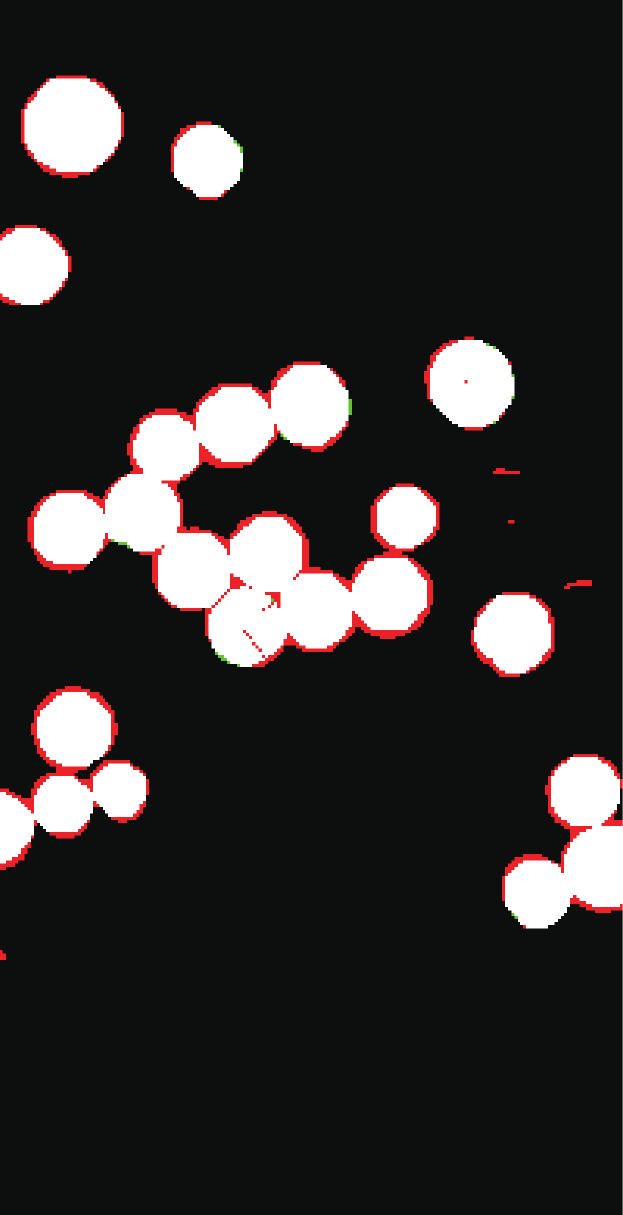}
        \label{fig:Her_MSDFFN}
    }
    \subfigure[]{
        \includegraphics[width=0.106\linewidth]{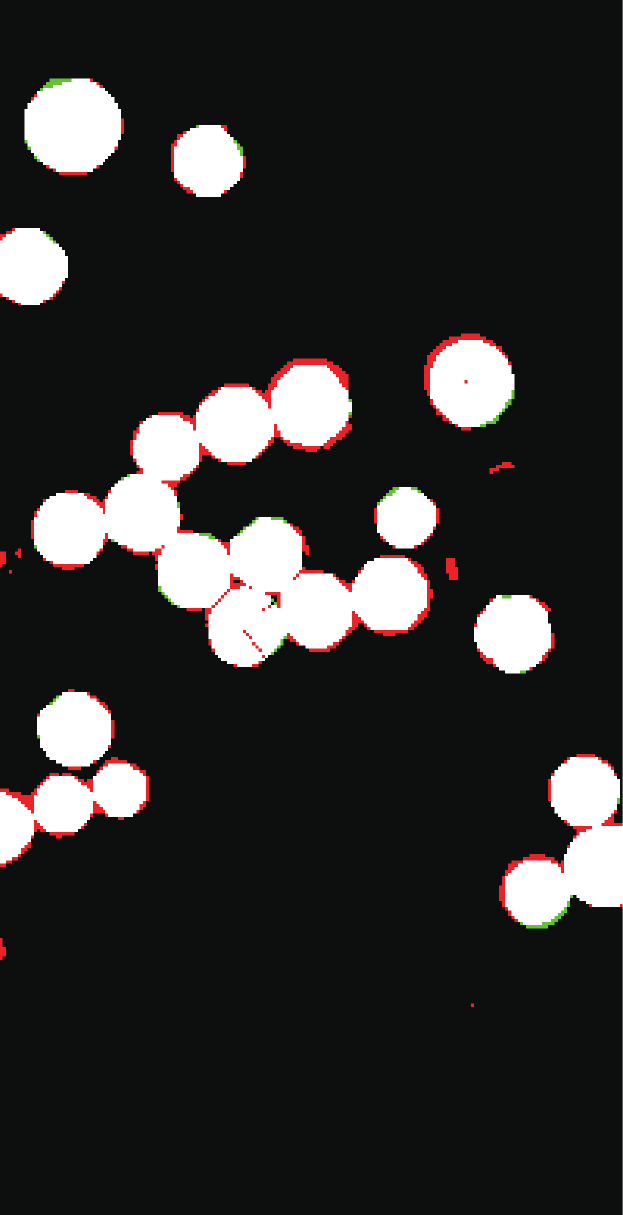}
        \label{fig:Her_HyGSTAN}
    }
    \subfigure[]{
        \includegraphics[width=0.106\linewidth]{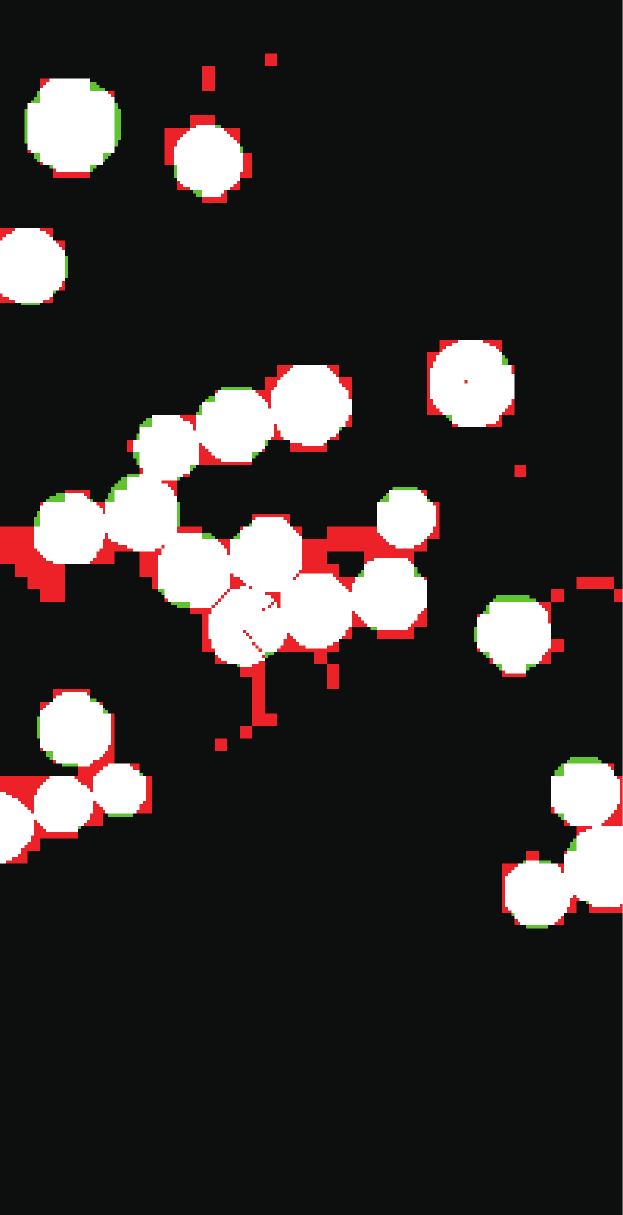}
        \label{fig:Her_G2AGCN}
    }
    \subfigure[]{
        \includegraphics[width=0.106\linewidth]{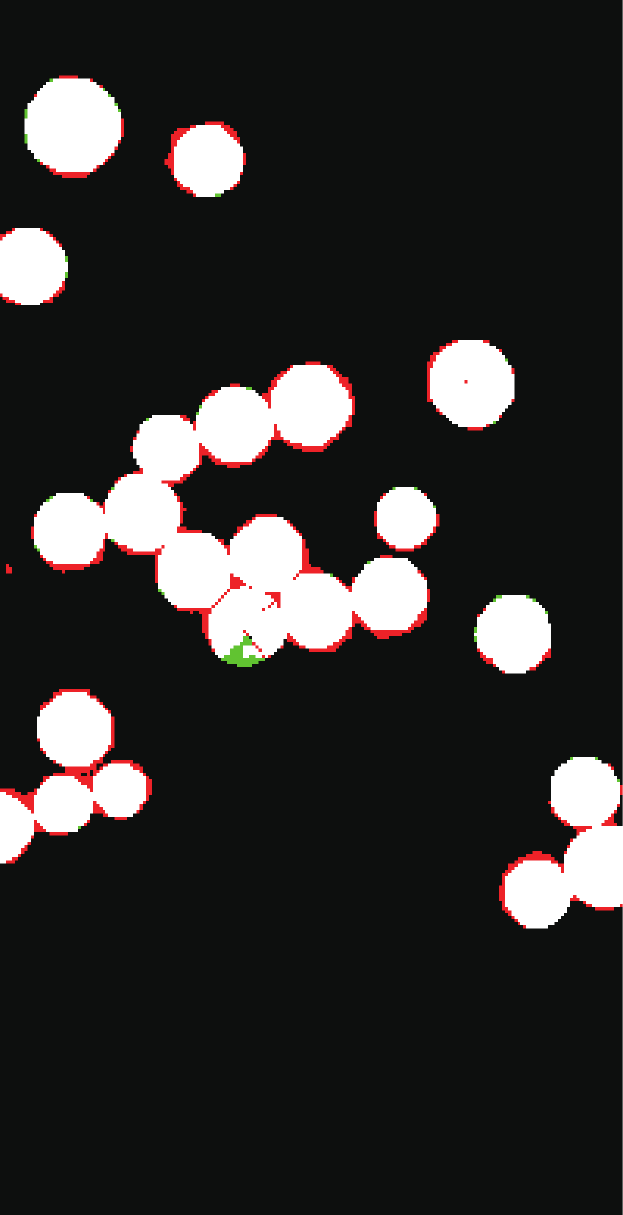}
        \label{fig:Her_CSDBF}
    }
    \subfigure[]{
        \includegraphics[width=0.106\linewidth]{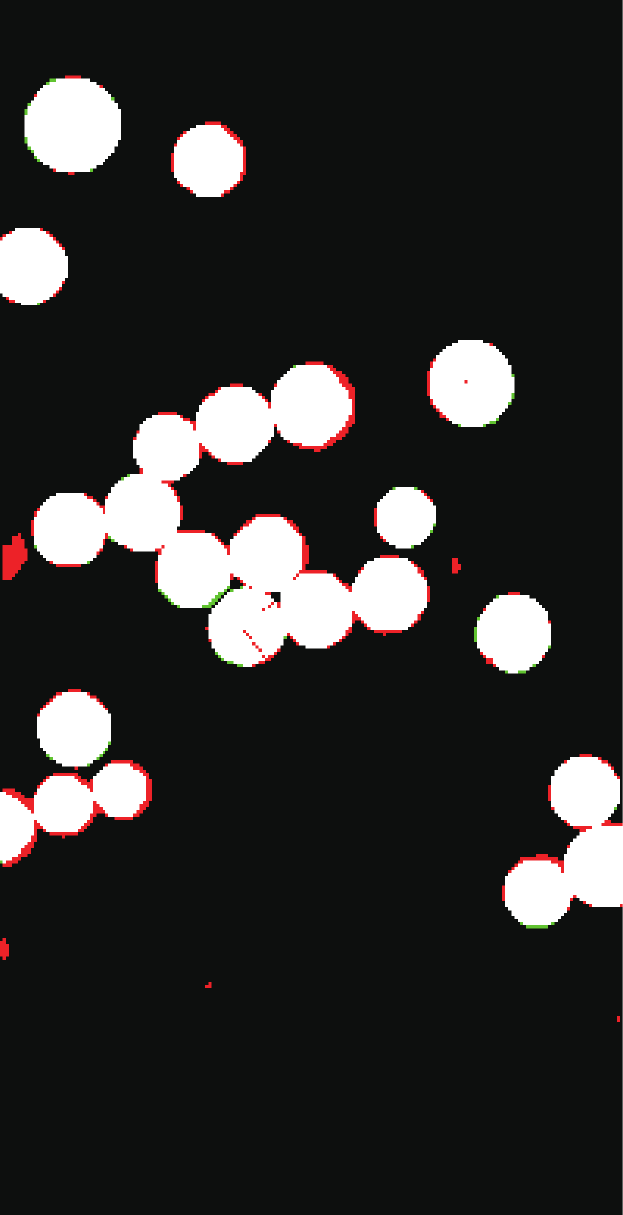}
        \label{fig:Her_proposed}
    }
    \caption{Change detection results of the HCD algorithms on the Hermiston dataset with FP and FN marked in red and green, respectively. (a) Ground Truth. (b) ML-EDAN. (c) SST-Former. (d) MSDFFN. (e) HyGSTAN. (f) $\textnormal{D}^2$AGCN. (g) CSDBF. (h) QUEEN-$\mathcal{G}$.}
    \label{fig:Her-exp}
    \vspace{-0.3cm}
\end{figure*}

\begin{figure*}[t]
    \centering
    \subfigure[]{
        \includegraphics[width=0.106\linewidth]{final_fig/Jia_GT.pdf}
        \label{fig:Jia_GT}
    }
    \subfigure[]{
        \includegraphics[width=0.106\linewidth]{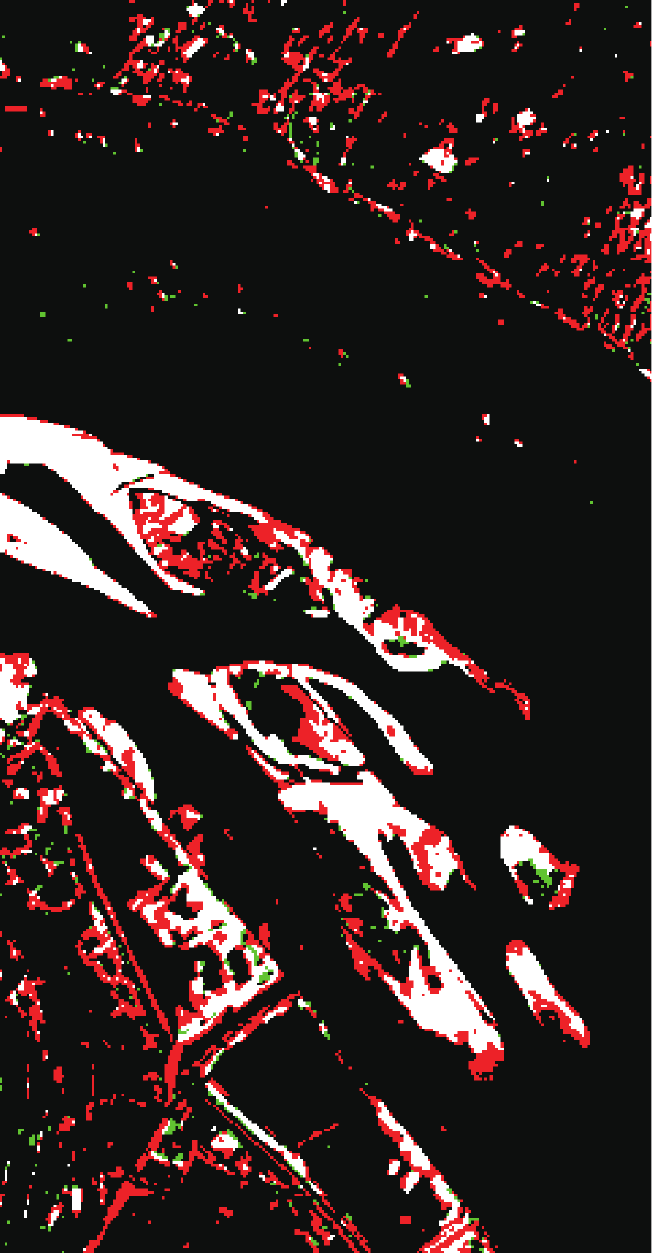}
        \label{fig:Jia_MLEDAN}
    }
    \subfigure[]{
        \includegraphics[width=0.106\linewidth]{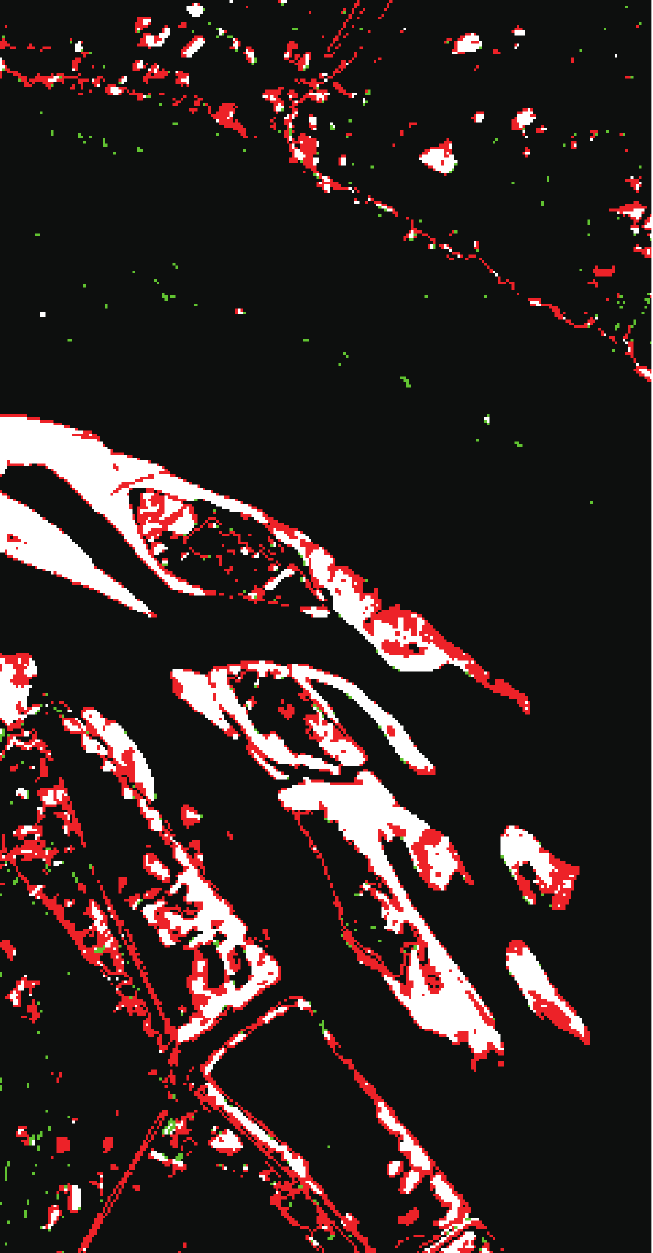}
        \label{fig:Jia_SSTF}
    }
    \subfigure[]{
        \includegraphics[width=0.106\linewidth]{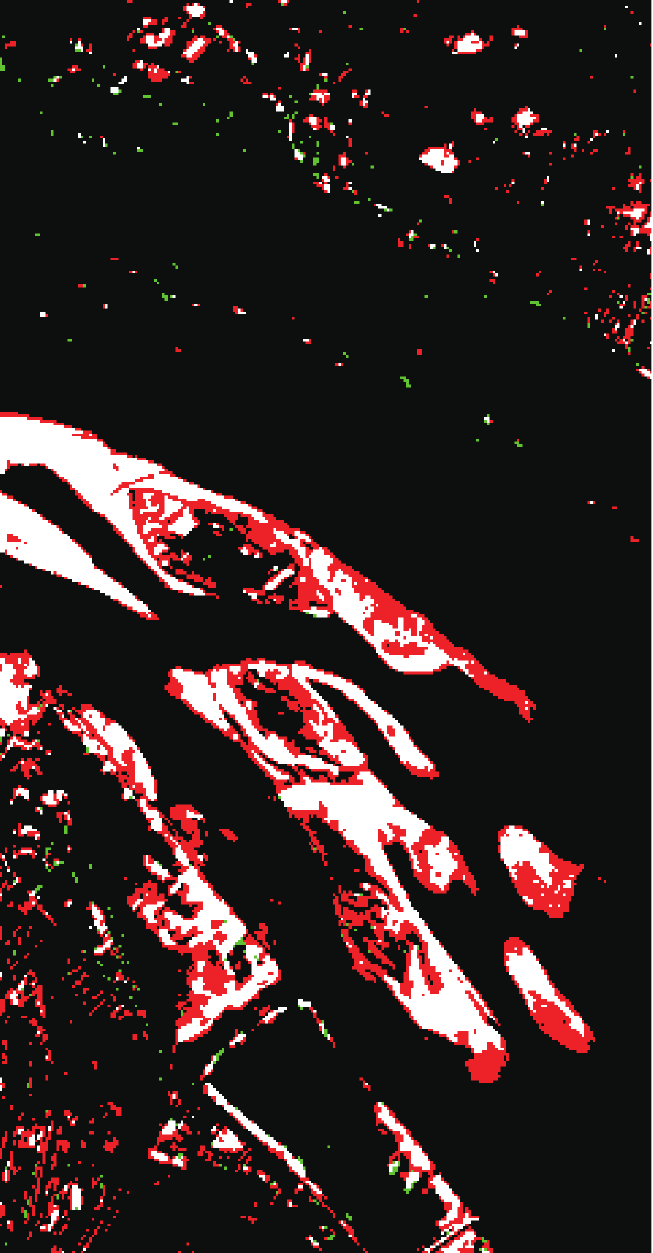}
        \label{fig:Jia_MSDFFN}
    }
    \subfigure[]{
        \includegraphics[width=0.106\linewidth]{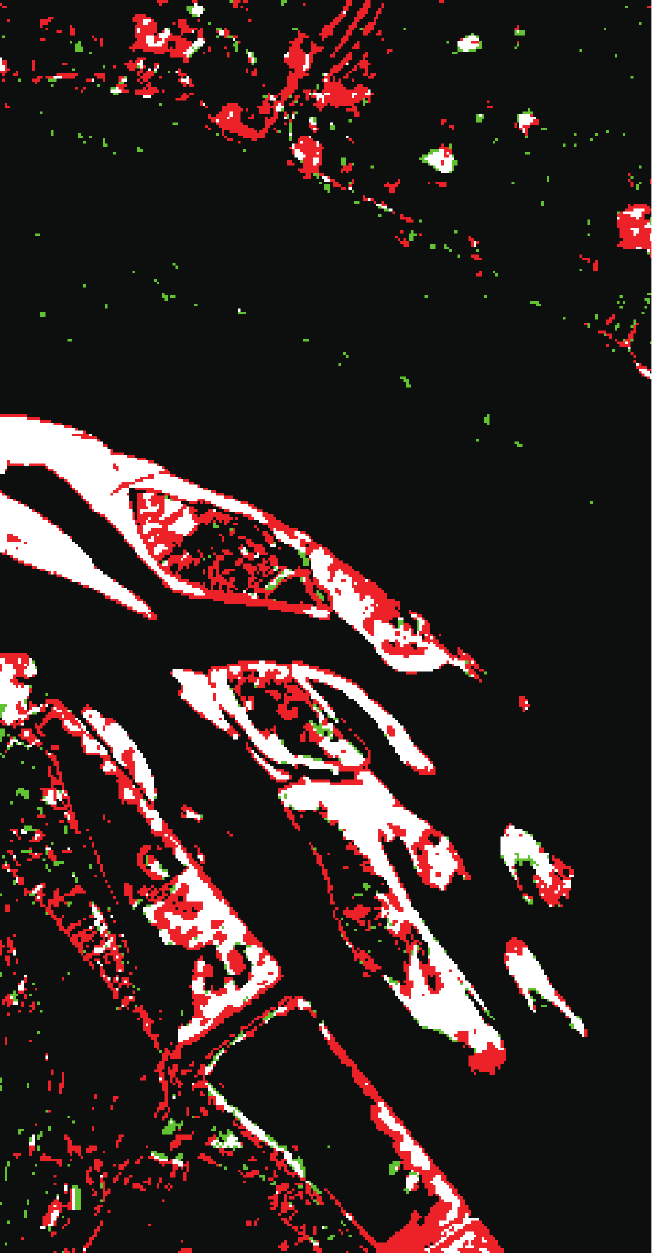}
        \label{fig:Jia_HyGSTAN}
    }
    \subfigure[]{
        \includegraphics[width=0.106\linewidth]{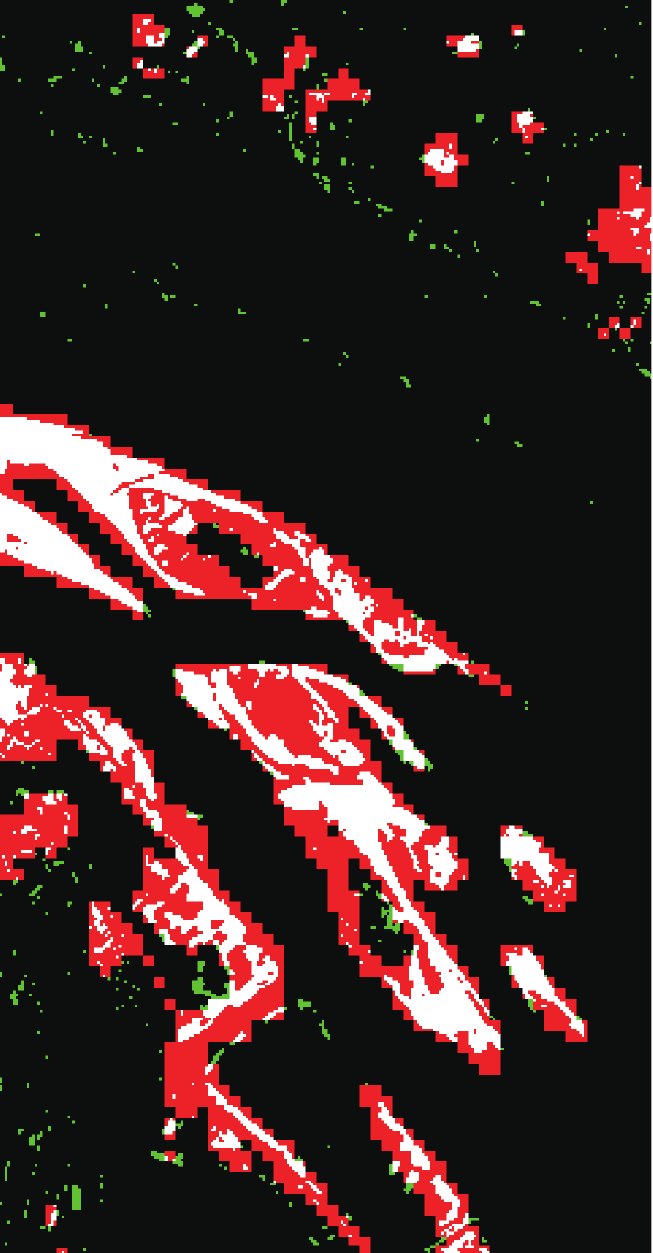}
        \label{fig:Jia_G2AGCN}
    }
    \subfigure[]{
        \includegraphics[width=0.106\linewidth]{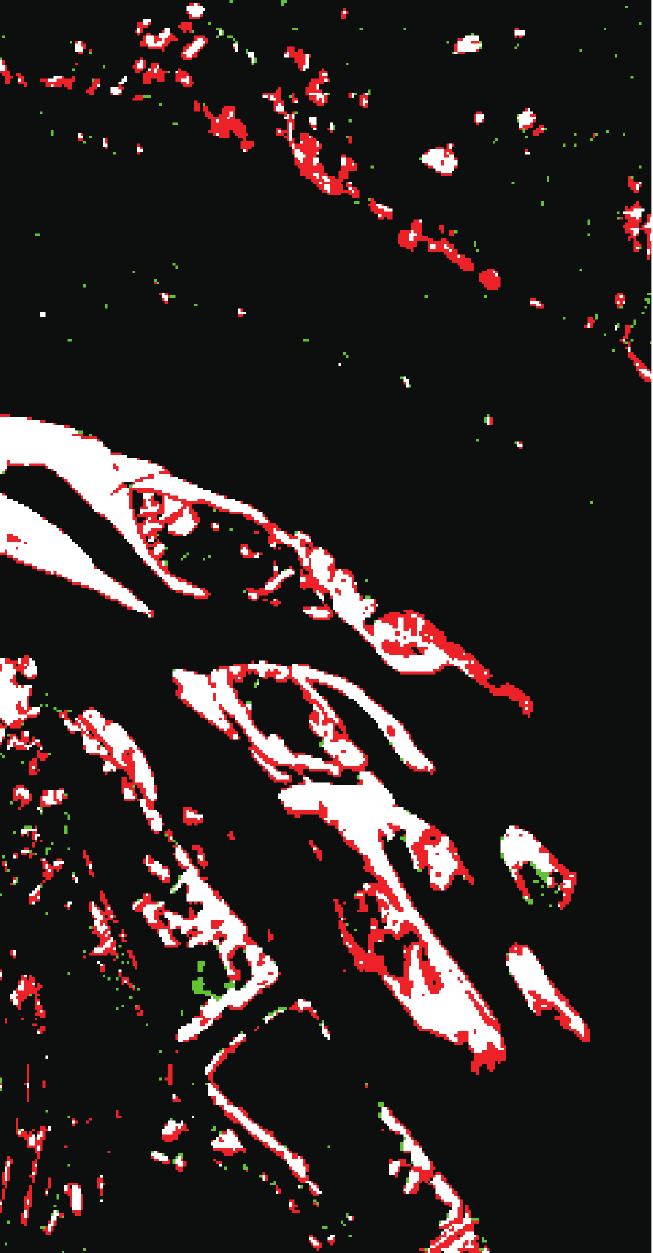}
        \label{fig:Jia_CSDBF}
    }
    \subfigure[]{
        \includegraphics[width=0.106\linewidth]{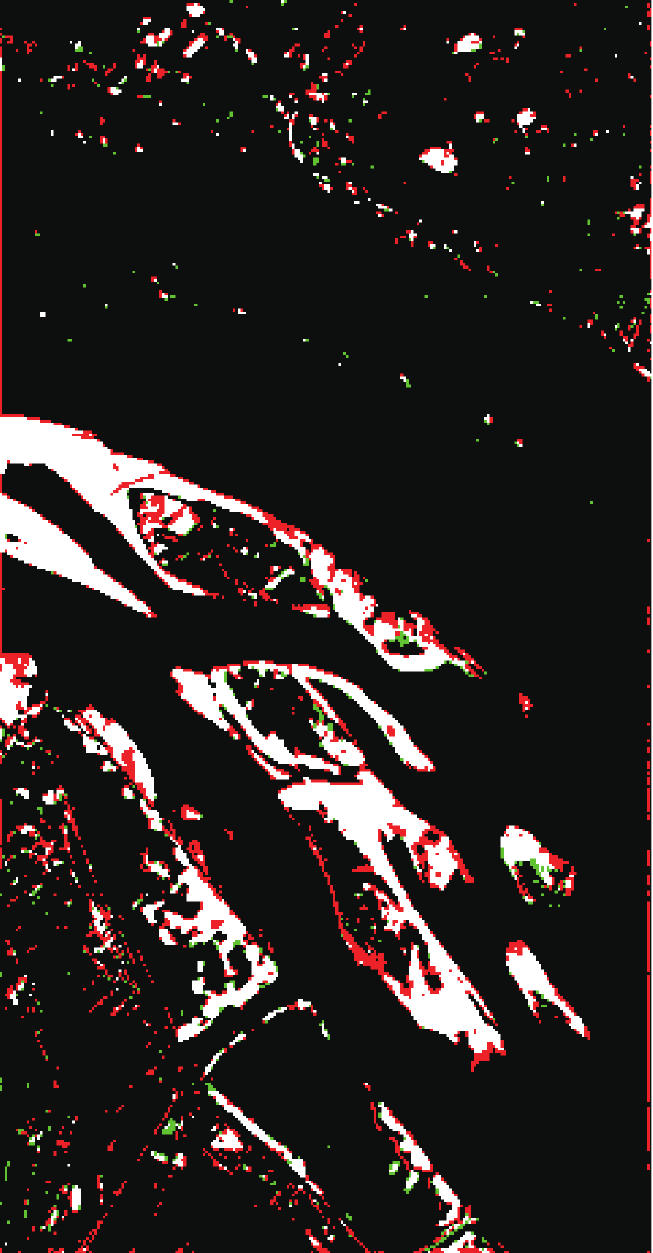}
        \label{fig:Jia_proposed}
    }
    \caption{Change detection results of the HCD algorithms on the Jiangsu dataset with FP and FN marked in red and green, respectively. (a) Ground Truth. (b) ML-EDAN. (c) SST-Former. (d) MSDFFN. (e) HyGSTAN. (f) $\textnormal{D}^2$AGCN. (g) CSDBF. (h) QUEEN-$\mathcal{G}$.}
    \label{fig:Jia-exp}
    \vspace{-0.3cm}
\end{figure*}

In this section, the quantitative and qualitative analysis of the proposed QUEEN-$\mathcal{G}$ and the benchmark methods, tested on the three widely used hyperspectral datasets, are detailed.
The quantitative result with five representative quantitative metrics is summarized in Table \ref{tab1:result}, with the best performance marked in an \textit{underlined boldfaced number}, while the second marked just using a \textit{boldfaced number}.
At the same time, the qualitative results are illustrated in Figs. \ref{fig:Yan-exp}-\ref{fig:Jia-exp}, with FP and FN marked in red and green, respectively.

\subsubsection{Experiment on the Yancheng Dataset}

From Table \ref{tab1:result}, the proposed QUEEN-$\mathcal{G}$ has the best performance on all three comprehensive indicators, i.e., OA, $\kappa$, and F1.
Moreover, the proposed method demonstrates a better balance between generating FN and FP in terms of the Pr and Re indicators, achieving the highest score in Pr.
The peer methods that achieve higher Pr or Re often do not perform as well on the other metric.
For instance, while MSDFFN achieves the best Re with a score of 0.979, its Pr is only 0.897.
In contrast, QUEEN-$\mathcal{G}$ demonstrates a more balanced performance, with a Re of 0.970 and a Pr of 0.926.
From the qualitative analysis of the Yancheng dataset shown in Fig. \ref{fig:Yan-exp}, QUEEN-$\mathcal{G}$ can detect the detail of the change areas at the farms in the middle, as shown in Fig. \ref{fig:Yan_proposed} , while other methods do not perform well at the details, e.g., MSDFFN, HyGSTAN, and $\textnormal{D}^2$AGCN in Figs. \ref{fig:Her_MSDFFN}-\ref{fig:Yan_G2AGCN}.
It is also worth noticing that the benchmark methods except MSDFFN all generate a large area of FP at the upper right of the change maps.
Combining both quantitative and qualitative analyses, the proposed QUEEN-$\mathcal{G}$ achieves the best performance on the Yancheng dataset.

\subsubsection{Experiment on the Hermiston Dataset}

As shown in Table \ref{tab1:result}, QUEEN-$\mathcal{G}$ achieves the best performance on the three comprehensive indicators while maintaining a good balance on the Pr and Re indices, achieving the highest and the second-highest scores in Pr and Re with 0.905 and 0.990, respectively.
Although MSDFFN achieves the highest Re score of 0.993, it only attains a Pr score of 0.887.
The result of the qualitative analysis also supports the observation of the quantitative analysis.
The peer methods struggle to clearly detect the changed pixels at the edge of the changed areas, for example, MSDFFN and HyGSTAN in Figs. \ref{fig:Her_MSDFFN} and \ref{fig:Her_HyGSTAN}.
Moreover, some of them generate considerable FP in the change maps, such as ML-EDAN, SST-Former, and $\textnormal{D}^2$AGCN in Figs. \ref{fig:Her_MLEDAN},  \ref{fig:Her_SSTF}and \ref{fig:Her_G2AGCN}.
CSDBF manages to control the generation of FPs; however, there is a region of FN in the middle of its detection map in Fig. \ref{fig:Her_CSDBF}.
Though there is a small number of regions of FP in the detection map of the proposed method, as shown in Fig. \ref{fig:Her_proposed}, overall, the number of FP and FN remains better than that of the benchmark methods.
QUEEN-$\mathcal{G}$ has the best performance on the Hermiston dataset in terms of quantitative and qualitative analyses.

\subsubsection{Experiment on the Jiangsu Dataset}

As can be observed from Table \ref{tab1:result}, the proposed QUEEN-$\mathcal{G}$ outperforms the peer methods in all three comprehensive indicators, maintaining a balanced performance on both the Pr and Re indices.
Notably, it also achieves the highest score in Pr with 0.608.
From Fig. \ref{fig:Jia-exp}, we can easily observe that the proposed method generates the fewest FPs in the entire detection map.
The peer methods struggle to correctly classify the scattered changed pixels at the upper (c.f. Figs. \ref{fig:Jia_HyGSTAN}-\ref{fig:Jia_CSDBF}) and the bottom-left of the images (c.f. Figs. \ref{fig:Jia_MLEDAN}-\ref{fig:Jia_G2AGCN}).
In comparing all GNN-based methods, one can observe that $\textnormal{D}^2$AGCN, which relies solely on GNN modules for learning superpixel-level features, exhibits limited capability in detecting scattered changed pixels, as illustrated in Fig. \ref{fig:Jia_G2AGCN}.
In contrast, CSDBF not only leverages GNN modules to learn features at the superpixel level but also integrates pixel-level information, resulting in a better performance compared to $\textnormal{D}^2$AGCN, as demonstrated in Fig. \ref{fig:Jia_CSDBF}.
The proposed QUEEN-$\mathcal{G}$ achieves the best performance among the GNN-based methods by utilizing a QUEEN-based module to enhance pixel-level information, which complements the GNN module, and by applying quantum unitary-computing information to the final detection stage, resulting in state-of-the-art detection performance, as shown in Fig. \ref{fig:Jia_proposed}.
Considering both the quantitative and qualitative analyses, QUEEN-$\mathcal{G}$ demonstrates superior performance over the benchmark methods on the Jiangsu dataset.

Combining the results of the above experiments, the proposed QUEEN-$\mathcal{G}$ demonstrates the best performance across all three datasets, in terms of both quantitative and qualitative analyses, compared to state-of-the-art HCD methods.
This shows the superiority of the proposed QUEEN-empowered graph neural network and the potential of introducing quantum computing to solve the HCD problem.

\subsection{Discussions of QUEEN-$\mathcal{G}$}\label{sec:discussion} 

In this section, we conduct a comprehensive analysis of the proposed QUEEN-$\mathcal{G}$ to evaluate its effectiveness and robustness. 
This analysis includes an ablation study that examines the contributions of the QFL module and the QEC module in Section \ref{sec:ablation}.
By isolating these components, we aim to understand their roles and contributions to the overall performance of the model. 
Furthermore, we explore the impact of varying sampling rates on QUEEN-$\mathcal{G}$ in Section \ref{sec:rate}, discussing how the proportion of sampled pixels influences the detection accuracy and stability across different datasets.

\subsubsection{Ablation Study}\label{sec:ablation}

\begingroup
\setlength{\tabcolsep}{6pt} 
\renewcommand{\arraystretch}{1.2} 
\begin{table}[t]
\begin{center}
\caption{Ablation study on the Yancheng and Hermiston datasets in terms of OA, $\kappa$, and F1 indices.} \label{tab:ablation}
\begin{tabular}{c|c|c|c|c|c|c|c}
\hline
\hline
\multicolumn{2}{c|}{Modules} &\multicolumn{3}{c|}{Yancheng}&\multicolumn{3}{c}{Hermiston}\\
\hline
QFL 
 & QEC & OA($\uparrow$) & $\kappa$($\uparrow$) & F1($\uparrow$)& OA($\uparrow$) & $\kappa$($\uparrow$) & F1($\uparrow$)\\
\hline
 &  & 0.966 & 0.917 & 0.941 & 0.985 & 0.932 & 0.942\\
\ding{51} &  & 0.967 & 0.920 & 0.944 & 0.984 & 0.932 & 0.941\\
 & \ding{51} & 0.967 & 0.921 & 0.944 & 0.983 & 0.928 & 0.937\\
\ding{51} & \ding{51} & {\bf 0.969} & {\bf 0.925} & {\bf 0.947} & {\bf 0.986} & {\bf 0.937} & {\bf 0.945}\\
\hline
\hline
\end{tabular}
\end{center}
\end{table}   
\endgroup

In this section, we investigate the effectiveness of the proposed quantum modules, i.e., the QFL module and the QEC classifier.
The QFL module contributes pixel-level unitary-computing features, while the QEC enhances classification capability by combining quantum and traditional classifiers, improving the effectiveness of the proposed QUEEN-$\mathcal{G}$.
For the QFL module, we evaluate the model's performance with and without the inclusion of QFL to assess the impact of quantum unitary-computing features at the pixel level.
Regarding the QEC, we examine the effectiveness of incorporating the QUEEN with FE into the classifier.
In the model without QEC, the final classification stage relies solely on the detection map generated by the traditional classifier, i.e., $\bM_\textnormal{FCL}$, as the only output.
The models are tested on the Yancheng and Hermiston datasets, with the results summarized in Table \ref{tab:ablation} based on OA, $\kappa$, and F1 indices.

In Table \ref{tab:ablation}, a checkmark indicates that the corresponding module is utilized within the model, whereas the absence of a checkmark denotes that the module is not employed.
The results presented in the second row of Table \ref{tab:ablation} demonstrate that individual inclusion of the QFL module enhances performance on the Yancheng dataset, though it slightly reduces performance on the Hermiston dataset compared to the baseline model shown in the first row.
Similarly, the third row indicates that individually incorporating the QEC module yields improved results on the Yancheng dataset but results in a minor decrease in performance on the Hermiston dataset relative to the baseline.
However, as shown in the fourth row, the combined inclusion of both QFL and QEC modules leads to the best performance across all metrics for both datasets.
For instance, in terms of $\kappa$, the model achieves an improvement of 0.008 over the baseline model of 0.917 on the Yancheng dataset and a 0.005 increase over the baseline model of 0.932 on the Hermiston dataset.
The experiment result shows that incorporating both QFL and QEC into the proposed method can generate the most accurate detection result with the proposed framework.

\subsubsection{Effect of Sampling Rate}\label{sec:rate}

\begin{figure*}[t]
\centering
    \subfigure[Yancheng]{
        \includegraphics[width=.31\linewidth]{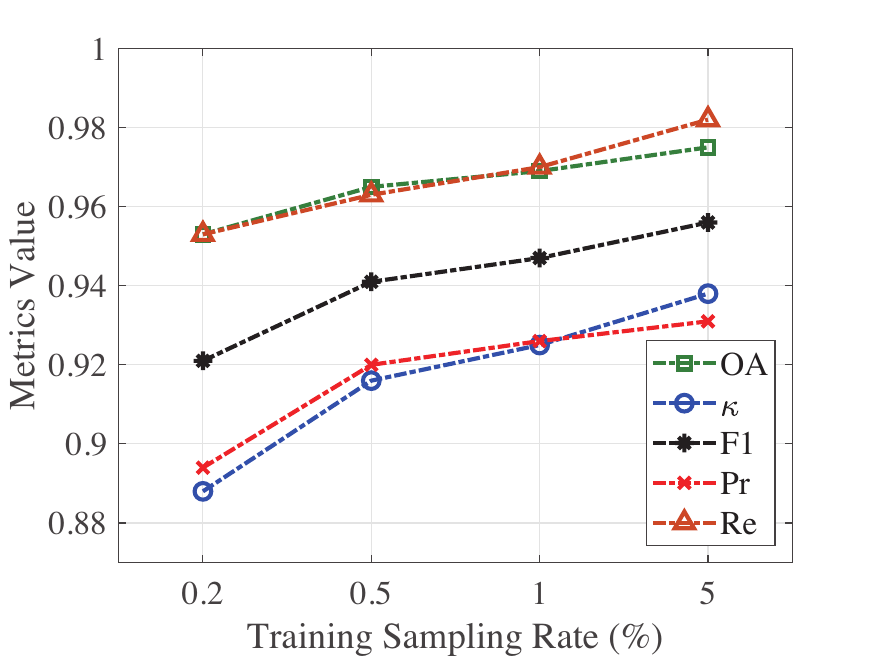}
        \label{fig:Yan_sam}
    }
    \subfigure[Hermiston]{
        \includegraphics[width=.31\linewidth]{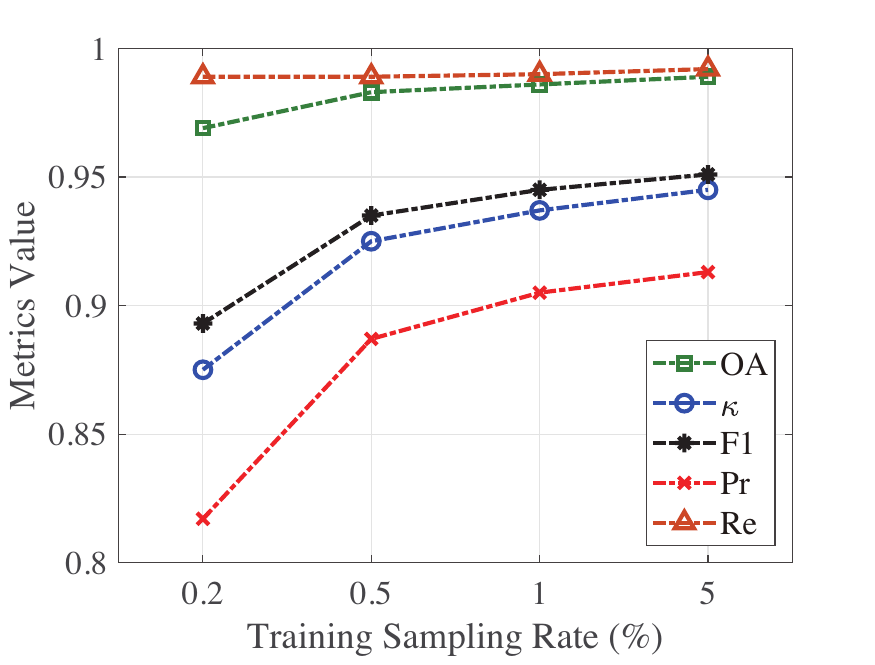}
        \label{fig:Her_sam}
    }
    \subfigure[Jiangsu]{
        \includegraphics[width=.31\linewidth]{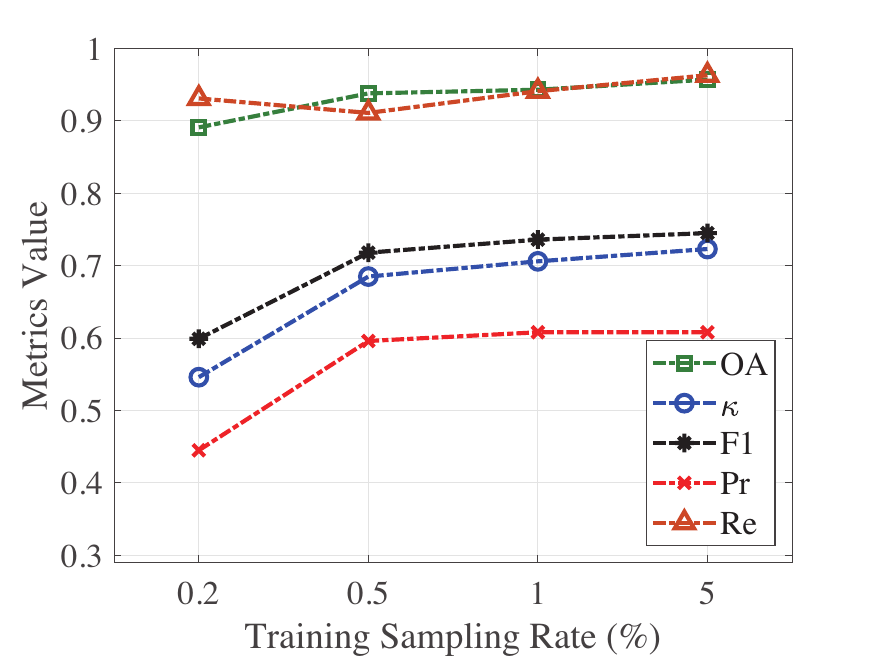}
        \label{fig:Jia_sam}
    }
    \caption{Performance comparison for different sampling rates (i.e., labeling rates) tested on the proposed QUEEN-$\mathcal{G}$ using three real HSI datasets.}\label{fig:exp2}
\end{figure*}

In this section, we examine the impact of the sampling rate on the performance of the proposed QUEEN-$\mathcal{G}$.
We perform the experiment on the three hyperspectral datasets with sampling rates ranging from 0.2\% to 5\%.
The experimental setup follows the procedure outlined in Section \ref{sec:setting}, with the only variation being the different sampling rates.
The result in terms of the five quantitative indices is illustrated in Fig. \ref{fig:exp2}.

Fig. \ref{fig:exp2} illustrates that the performance proposed QUEEN-$\mathcal{G}$ generally improves across all five indices as the sampling rate increases.
This trend is particularly evident in the Yancheng dataset shown in Fig. \ref{fig:Yan_sam}, where higher sampling rates consistently lead to enhanced outcomes.
However, for the Hermiston dataset, illustrated in Fig. \ref{fig:Her_sam}, although the Re values do not increase with the sampling rate, the Pr values show significant improvement, leading to better overall performance as reflected in the comprehensive indices of OA, $\kappa$, and F1.
For the Jiangsu dataset, the performance of the model stabilizes when the sampling rate exceeds $0.5\%$ as shown in Fig. \ref{fig:Jia_sam}, indicating that QUEEN-$\mathcal{G}$ maintains robust and reliable performance even at a lower sampling rate.
This stability across varying sampling rates demonstrates the resilience and adaptability of the proposed method, ensuring consistent performance even when data availability is limited.

\subsubsection{Parameter Sensitivity Analysis}\label{sec:parameter}

\begin{figure*}[t]
\centering
    \subfigure[Yancheng]{
        \includegraphics[width=.31\linewidth]{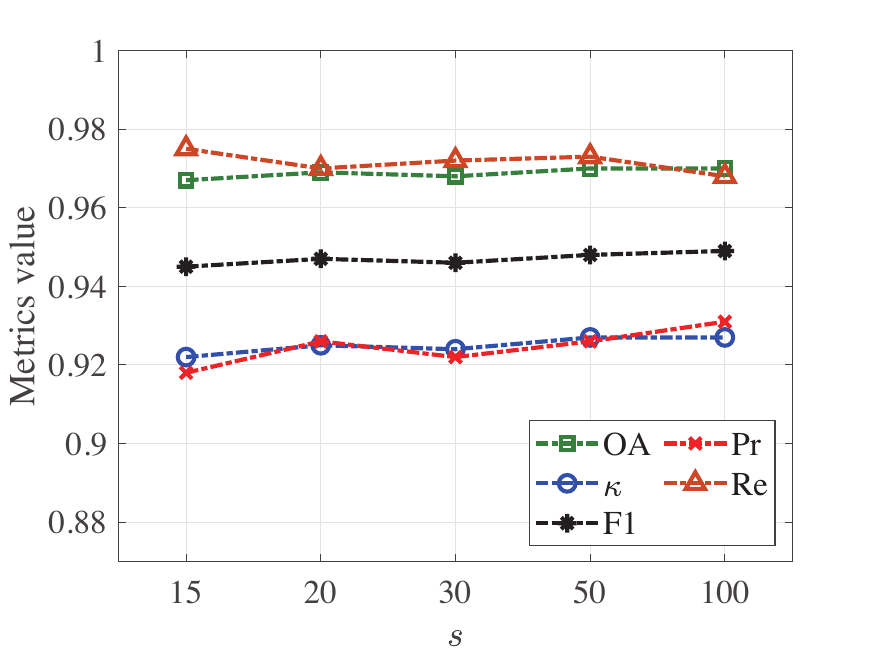}
        \label{fig:Yan_sen}
    }
    \subfigure[Hermiston]{
        \includegraphics[width=.31\linewidth]{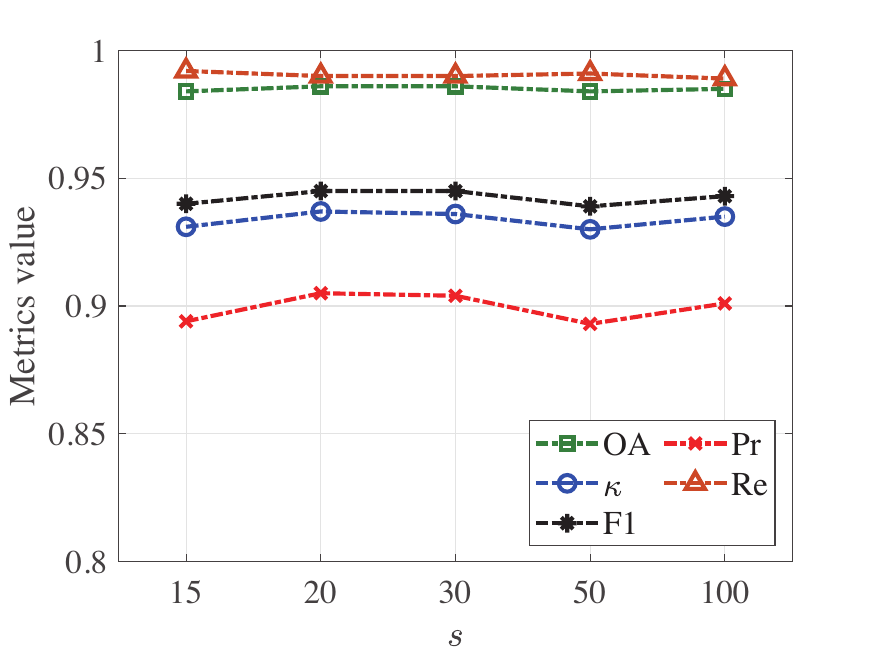}
        \label{fig:Her_sen}
    }
    \subfigure[Jiangsu]{
        \includegraphics[width=.31\linewidth]{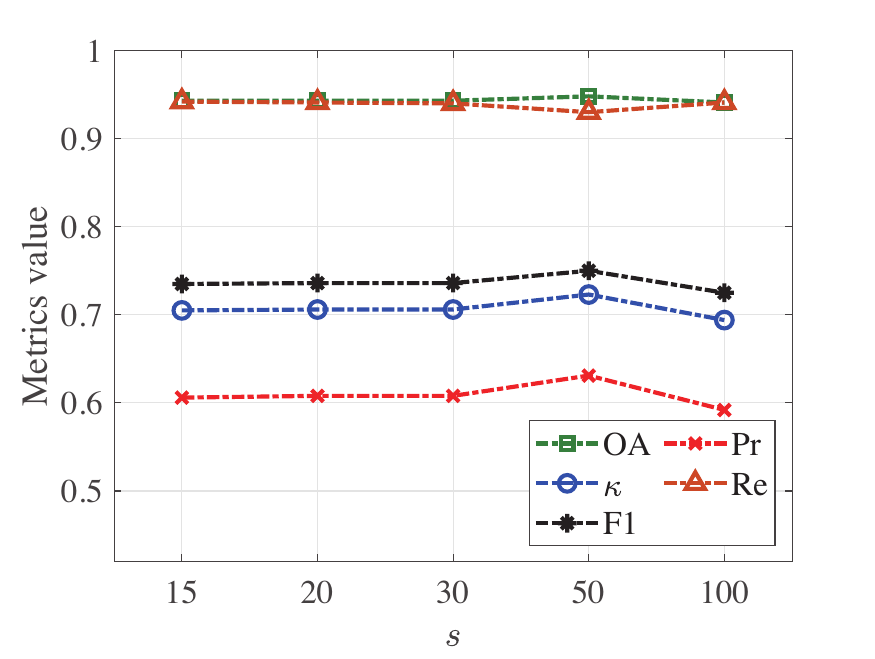}
        \label{fig:Jia_sen}
    }
    \caption{Performance comparison for different parameter values of the superpixel segmentation scale $s$ in QUEEN-$\mathcal{G}$ using three real HSI datasets.}\label{fig:par_sen}
\end{figure*}

In this section, we investigate the effect of the key parameter, i.e., the superpixel segmentation scale $s$ for constructing the graph in GFL.
We perform the experiment on the three hyperspectral datasets by varying $s$ over a wide range of values, i.e., $s \in \{15, 20, 30, 50, 100 \}$.
The results in terms of the five quantitative indices is illustrated in Fig. \ref{fig:par_sen}.
For the Yancheng dataset, the optimal performance is achieved when $s:=100$.
For the Hermiston dataset, the optimal performance is achieved when $s:=20$.
For the Jiangsu dataset, the optimal performance is achieved when $s:=50$.
These findings suggest that fine-tuning the parameters for each dataset has the potential to further enhance the performance of QUEEN-$\mathcal{G}$.
However, as can be observed from Fig. \ref{fig:par_sen}, QUEEN-$\mathcal{G}$ is generally not sensitive to the parameter setting of $s$, allowing us to adopt a unified setting of $s:=20$, which still ensures stable performances for different data and different metrics.
This makes QUEEN-$\mathcal{G}$ more practical and user-friendly.

\begingroup
\setlength{\tabcolsep}{10pt} 
\renewcommand{\arraystretch}{1.2} 
\begin{table}[t]
\begin{center}
\caption{Comparison of computational efficiency measured in running time (in seconds) for benchmark HCD algorithms and the proposed QUEEN-$\mathcal{G}$.} \label{tab:time}
	\scalebox{1.1}{
\begin{tabular}{c|c|c|c}
\hline
\hline
Methods & Yancheng & Hermiston & Jiangsu\\
\hline
ML-EDAN & 354.34 & 415.15 & 563.93\\
SST-Former & 141.20 & 182.00 & 247.73\\
MSDFFN & 209.19 & 255.58 & 340.93\\
HyGSTAN & 5.29 & 6.35 & 6.37\\
$\textnormal{D}^2$AGCN & 1154.95 & 1472.89 & 2207.24 \\
CSDBF & 124.69 & 187.87 & 324.42 \\
QUEEN-$\mathcal{G}$ & 206.09 & 253.64 & 360.29 \\
\hline
\hline

\end{tabular}}
\end{center}
\vspace{-0.25cm}
\end{table}   
\endgroup

\subsubsection{Computational Efficiency Analysis}\label{sec:computational}
Finally, we analyze the computational efficiency of the proposed QUEEN-$\mathcal{G}$ algorithm.
From the mathematical definitions of the quantum neurons in Table \ref{tab:common_qu_gate} and some basic properties of Kronecker products, one can derive the computational complexities for the two quantum modules (i.e., QFL and QEC), both of which are $\mathcal{O}\left(HW \times \left(\eta \times 2\times 2^{2q}+\lambda \times 2\times (2^{2q}+2^{q})\right)\right)$ with $q$ being the number of qubits, $\eta$ being the number of quantum layers, $\lambda$ being the number of quantum measurements, and $H\times W$ being the number of pixels.
On the other hand, the values of floating point operations (FLOPs) for the quantum part in QUEEN-$\mathcal{G}$ (i.e., QFL and QEC) are 1173.31K for the Yancheng data, 1452.67K for the Hermiston data, and 2078.12K for the Jiangsu data, respectively.
Together with the non-quantum parts, the values of FLOPs for the entire QUEEN-$\mathcal{G}$ are 10.68G for the Yancheng data, 13.16G for the Hermiston data, and 18.55G for the Jiangsu data, respectively.
To get a sense about the complexity, we further measure the computational efficiency as the running time (in seconds).
From Table \ref{tab:time}, one can see that QUEEN-$\mathcal{G}$ is the fourth fastest among the seven HCD methods under test.
For all these benchmark hyperspectral datasets, the proposed QUEEN-$\mathcal{G}$ algorithm can complete the detection procedure in just a few minutes.
It can also be observed that as the total number of pixels in the bi-temporal HSIs increases, the computational time also increases for all the HCD methods, where HyGSTAN has the fastest computational time with a compromise of the HCD performance, as shown in Section \ref{sec:performance}.

To summarize Section \ref{sec:experiment}, the experimental findings underscore the effectiveness and robustness of the proposed method. 
Integrating both the QUEEN-based modules, i.e., QFL and QEC, within the framework yields the best detection results, showcasing the superior capability of the proposed QUEEN-$\mathcal{G}$.
Moreover, the consistent performance across varying sampling rates highlights the resilience and adaptability of QUEEN-$\mathcal{G}$, affirming its reliability even in scenarios with limited data availability.
Furthermore, the unified parameter setting is proved to yield stable performance across all three real hyperspectral datasets.
Overall, the proposed QUEEN-$\mathcal{G}$ strikes a good balance between the computational efficiency and the detection performance, leading to state-of-the-art HCD results with decent computational time.

\section{Conclusion and Future Works}\label{sec:conclusion}

For the first time, this work introduces quantum computing into the critical change detection task in hyperspectral remote sensing.
This allows us to obtain radically new quantum unitary-computing features, serving as new information for better decision-making in the final detection stage.
We employ the graph neural network (GNN) to learn the graph features at the superpixel level, while the quantum deep network (QUEEN) learns the quantum features at the pixel level for obtaining finer spatial information not retained by GNN.
Both features are fused and fed into a quantum classifier to make the final decision of whether there is a change or not.
The adopted QUEEN architecture has the mathematically provable FE, meaning that it has the capability to express any valid quantum function, thereby allowing the effective extraction of the entangled pixel-level quantum features.
The proposed QUEEN-empowered GNN (i.e., QUEEN-$\mathcal{G}$) is trained using a customized loss function judiciously designed to ensure the discriminative capability of the classifier.
QUEEN-$\mathcal{G}$ achieves state-of-the-art HCD performances on real benchmark datasets, where ablation study demonstrates the critical role of the quantum features.

Despite the promising results achieved by the proposed method, the well-known quantum no-cloning theorem of quantum mechanics has imposed a fundamental limitation on the design of QUEEN-based modules, restricting their flexibility compared to conventional CNN modules.
For example, no-cloning theorem alludes that the effective structure of residual network (ResNet, requiring copying the feature information to some deeper layers) cannot be implemented under the quantum framework.
This also indicates the importance of complementary techniques to achieve advanced QNNs, as mentioned in \cite{lin2023hyperqueen}.
In the future, we aim at using quantum features as new information, to upgrade the performances of critical techniques and to explore more cutting-edge applications in the remote sensing areas and beyond.

\renewcommand{\thesubsection}{\Alph{subsection}}
\bibliography{ref}

\begin{thebibliography}{10}
\providecommand{\url}[1]{#1}
\csname url@samestyle\endcsname
\providecommand{\newblock}{\relax}
\providecommand{\bibinfo}[2]{#2}
\providecommand{\BIBentrySTDinterwordspacing}{\spaceskip=0pt\relax}
\providecommand{\BIBentryALTinterwordstretchfactor}{4}
\providecommand{\BIBentryALTinterwordspacing}{\spaceskip=\fontdimen2\font plus
\BIBentryALTinterwordstretchfactor\fontdimen3\font minus
  \fontdimen4\font\relax}
\providecommand{\BIBforeignlanguage}[2]{{%
\expandafter\ifx\csname l@#1\endcsname\relax
\typeout{** WARNING: IEEEtran.bst: No hyphenation pattern has been}%
\typeout{** loaded for the language `#1'. Using the pattern for}%
\typeout{** the default language instead.}%
\else
\language=\csname l@#1\endcsname
\fi
#2}}
\providecommand{\BIBdecl}{\relax}
\BIBdecl

\bibitem{abd2011land}
O.~Abd El-Kawy, J.~R{\o}d, H.~Ismail, and A.~Suliman, ``Land use and land cover
  change detection in the western nile delta of egypt using remote sensing
  data,'' \emph{Applied Geography}, vol.~31, no.~2, pp. 483--494, Apr. 2011.

\bibitem{bovolo2007split}
F.~Bovolo and L.~Bruzzone, ``A split-based approach to unsupervised change
  detection in large-size multitemporal images: Application to tsunami-damage
  assessment,'' \emph{IEEE Transactions on Geoscience and Remote Sensing},
  vol.~45, no.~6, pp. 1658--1670, Jun. 2007.

\bibitem{yang2003urban}
L.~Yang, G.~Xian, J.~M. Klaver, and B.~Deal, ``Urban land-cover change
  detection through sub-pixel imperviousness mapping using remotely sensed
  data,'' \emph{Photogrammetric Engineering \& Remote Sensing}, vol.~69, no.~9,
  pp. 1003--1010, Sep. 2003.

\bibitem{bioucas2013hyperspectral}
J.~M. Bioucas-Dias, A.~Plaza, G.~Camps-Valls, P.~Scheunders, N.~Nasrabadi, and
  J.~Chanussot, ``Hyperspectral remote sensing data analysis and future
  challenges,'' \emph{IEEE Geoscience and Remote Sensing Magazine}, vol.~1,
  no.~2, pp. 6--36, Aug. 2013.

\bibitem{li2024learning}
C.~Li, B.~Zhang, D.~Hong, X.~Jia, A.~Plaza, and J.~Chanussot, ``Learning
  disentangled priors for hyperspectral anomaly detection: A coupling
  model-driven and data-driven paradigm,'' \emph{IEEE Transactions on Neural
  Networks and Learning Systems}, pp. 1--14, 2024.

\bibitem{lin2021all}
C.-H. Lin and T.-H. Lin, ``All-addition hyperspectral compressed sensing for
  metasurface-driven miniaturized satellite,'' \emph{IEEE Transactions on
  Geoscience and Remote Sensing}, vol.~60, pp. 1--15, Mar. 2022.

\bibitem{lin2024signal}
C.-H. Lin and S.-S. Young, ``Signal subspace identification for incomplete
  hyperspectral image with applications to various inverse problems,''
  \emph{IEEE Transactions on Geoscience and Remote Sensing}, vol.~62, pp.
  1--16, Mar. 2024.

\bibitem{du2012fusion}
P.~Du, S.~Liu, P.~Gamba, K.~Tan, and J.~Xia, ``Fusion of difference images for
  change detection over urban areas,'' \emph{IEEE Journal of Selected Topics in
  Applied Earth Observations and Remote Sensing}, vol.~5, no.~4, pp.
  1076--1086, Aug. 2012.

\bibitem{falco2012change}
N.~Falco, M.~D. Mura, F.~Bovolo, J.~A. Benediktsson, and L.~Bruzzone, ``Change
  detection in {VHR} images based on morphological attribute profiles,''
  \emph{IEEE Geoscience and Remote Sensing Letters}, vol.~10, no.~3, pp.
  636--640, May 2013.

\bibitem{rignot1993change}
E.~J.~M. Rignot and J.~J. van Zyl, ``Change detection techniques for {ERS-1
  SAR} data,'' \emph{IEEE Transactions on Geoscience and Remote Sensing},
  vol.~31, no.~4, pp. 896--906, Jul. 1993.

\bibitem{zheng2019unsupervised}
W.-C. Zheng, C.-H. Lin, K.-H. Tseng, C.-Y. Huang, T.-H. Lin, C.-H. Wang, and
  C.-Y. Chi, ``Unsupervised change detection in multitemporal multispectral
  satellite images: A convex relaxation approach,'' in \emph{Proc. IEEE
  International Geoscience and Remote Sensing Symposium (IGARSS)}, Yokohama,
  Japan, Jul. 28-Aug. 02, 2019, pp. 1546--1549.

\bibitem{lin2015identifiability}
C.-H. Lin, W.-K. Ma, W.-C. Li, C.-Y. Chi, and A.~Ambikapathi, ``Identifiability
  of the simplex volume minimization criterion for blind hyperspectral
  unmixing: {The} no-pure-pixel case,'' \emph{IEEE Transactions on Geoscience
  and Remote Sensing}, vol.~53, no.~10, pp. 5530--5546, May 2015.

\bibitem{lin2018maximum}
C.-H. Lin, R.~Wu, W.-K. Ma, C.-Y. Chi, and Y.~Wang, ``Maximum volume inscribed
  ellipsoid: {A} new simplex-structured matrix factorization framework via
  facet enumeration and convex optimization,'' \emph{SIAM Journal on Imaging
  Sciences}, vol.~11, no.~2, pp. 1651--1679, Jun. 2018.

\bibitem{bovolo2006theoretical}
F.~Bovolo and L.~Bruzzone, ``A theoretical framework for unsupervised change
  detection based on change vector analysis in the polar domain,'' \emph{IEEE
  Transactions on Geoscience and Remote Sensing}, vol.~45, no.~1, pp. 218--236,
  Jan. 2007.

\bibitem{wu2013slow}
C.~Wu, B.~Du, and L.~Zhang, ``Slow feature analysis for change detection in
  multispectral imagery,'' \emph{IEEE Transactions on Geoscience and Remote
  Sensing}, vol.~52, no.~5, pp. 2858--2874, May 2014.

\bibitem{tewkesbury2015critical}
A.~P. Tewkesbury, A.~J. Comber, N.~J. Tate, A.~Lamb, and P.~F. Fisher, ``A
  critical synthesis of remotely sensed optical image change detection
  techniques,'' \emph{Remote Sensing of Environment}, vol. 160, pp. 1--14, Feb.
  2015.

\bibitem{nielsen1998multivariate}
A.~A. Nielsen, K.~Conradsen, and J.~J. Simpson, ``Multivariate alteration
  detection ({MAD}) and {MAF} postprocessing in multispectral, bitemporal image
  data: New approaches to change detection studies,'' \emph{Remote Sensing of
  Environment}, vol.~64, no.~1, pp. 1--19, Apr. 1998.

\bibitem{nemmour2006multiple}
H.~Nemmour and Y.~Chibani, ``Multiple support vector machines for land cover
  change detection: An application for mapping urban extensions,'' \emph{ISPRS
  Journal of Photogrammetry and Remote Sensing}, vol.~61, no.~2, pp. 125--133,
  Oct. 2006.

\bibitem{guo2016deep}
Y.~Guo, Y.~Liu, A.~Oerlemans, S.~Lao, S.~Wu, and M.~S. Lew, ``Deep learning for
  visual understanding: A review,'' \emph{Neurocomputing}, vol. 187, pp.
  27--48, Apr. 2016.

\bibitem{deng2014deep}
L.~Deng and D.~Yu, ``Deep learning: Methods and applications,''
  \emph{Foundations and Trends{\textregistered} in Signal Processing}, vol.~7,
  no. 3–4, pp. 197--387, Jun. 2014.

\bibitem{lin2024qrcode}
C.-H. Lin, C.-C. Hsu, S.-S. Young, C.-Y. Hsieh, and S.-C. Tai, ``{QRCODE}:
  Quasi-residual convex deep network for fusing misaligned hyperspectral and
  multispectral images,'' \emph{IEEE Transactions on Geoscience and Remote
  Sensing}, vol.~62, pp. 1--15, Mar. 2024.

\bibitem{dong2018local}
H.~Dong, W.~Ma, Y.~Wu, M.~Gong, and L.~Jiao, ``Local descriptor learning for
  change detection in synthetic aperture radar images via convolutional neural
  networks,'' \emph{IEEE Access}, vol.~7, pp. 15\,389--15\,403, Dec. 2019.

\bibitem{wang2018getnet}
Q.~Wang, Z.~Yuan, Q.~Du, and X.~Li, ``{GETNET}: A general end-to-end 2-{D}
  {CNN} framework for hyperspectral image change detection,'' \emph{IEEE
  Transactions on Geoscience and Remote Sensing}, vol.~57, no.~1, pp. 3--13,
  Jan. 2019.

\bibitem{lin2019multispectral}
Y.~Lin, S.~Li, L.~Fang, and P.~Ghamisi, ``Multispectral change detection with
  bilinear convolutional neural networks,'' \emph{IEEE Geoscience and Remote
  Sensing Letters}, vol.~17, no.~10, pp. 1757--1761, Oct. 2020.

\bibitem{qu2021multilevel}
J.~Qu, S.~Hou, W.~Dong, Y.~Li, and W.~Xie, ``A multilevel encoder–decoder
  attention network for change detection in hyperspectral images,'' \emph{IEEE
  Transactions on Geoscience and Remote Sensing}, vol.~60, pp. 1--13, Nov.
  2022.

\bibitem{tang2024transformer}
P.-W. Tang, C.-H. Lin, and Y.~Liu, ``Transformer-driven inverse problem
  transform for fast blind hyperspectral image dehazing,'' \emph{IEEE
  Transactions on Geoscience and Remote Sensing}, vol.~62, pp. 1--14, Jan.
  2024.

\bibitem{young2023cidar}
S.-S. Young, C.-H. Lin, and J.-T. Lin, ``{CiDAR-Former}: Cosine-weighting deep
  abundance reconstruction transformer for fast unsupervised hyperspectral
  anomaly detection,'' in \emph{Proc. 2023 13th Workshop on Hyperspectral
  Imaging and Signal Processing: Evolution in Remote Sensing (WHISPERS)},
  Athens, Greece, Oct. 31- Nov. 02, 2023, pp. 1--5.

\bibitem{song2022csanet}
R.~Song, W.~Ni, W.~Cheng, and X.~Wang, ``{CSANet}: Cross-temporal interaction
  symmetric attention network for hyperspectral image change detection,''
  \emph{IEEE Geoscience and Remote Sensing Letters}, vol.~19, pp. 1--5, May
  2022.

\bibitem{wang2022spectral}
Y.~Wang, D.~Hong, J.~Sha, L.~Gao, L.~Liu, Y.~Zhang, and X.~Rong,
  ``Spectral–spatial–temporal transformers for hyperspectral image change
  detection,'' \emph{IEEE Transactions on Geoscience and Remote Sensing},
  vol.~60, pp. 1--14, Aug. 2022.

\bibitem{luo2023multiscale}
F.~Luo, T.~Zhou, J.~Liu, T.~Guo, X.~Gong, and J.~Ren, ``Multiscale diff-changed
  feature fusion network for hyperspectral image change detection,'' \emph{IEEE
  Transactions on Geoscience and Remote Sensing}, vol.~61, pp. 1--13, Jan.
  2023.

\bibitem{yu2024hyperspectral}
H.~Yu, H.~Yang, L.~Gao, J.~Hu, A.~Plaza, and B.~Zhang, ``Hyperspectral image
  change detection based on gated spectral–spatial–temporal attention
  network with spectral similarity filtering,'' \emph{IEEE Transactions on
  Geoscience and Remote Sensing}, vol.~62, pp. 1--13, Mar. 2024.

\bibitem{hong2020graph}
D.~Hong, L.~Gao, J.~Yao, B.~Zhang, A.~Plaza, and J.~Chanussot, ``Graph
  convolutional networks for hyperspectral image classification,'' \emph{IEEE
  Transactions on Geoscience and Remote Sensing}, vol.~59, no.~7, pp.
  5966--5978, Jul. 2021.

\bibitem{yu2022unsupervised}
C.~Yu, S.~Zhou, M.~Song, B.~Gong, E.~Zhao, and C.-I. Chang, ``Unsupervised
  hyperspectral band selection via hybrid graph convolutional network,''
  \emph{IEEE Transactions on Geoscience and Remote Sensing}, vol.~60, pp.
  1--15, Jun. 2022.

\bibitem{lin2023gnn}
T.-H. Lin, C.-H. Lin, and S.-S. Young, ``{GNN}-based small-data learning with
  area-control mechanism for hyperspectral satellite change detection,'' in
  \emph{Proc. 2023 Asia Pacific Signal and Information Processing Association
  Annual Summit and Conference (APSIPA ASC)}, Taipei, Taiwan, Oct. 31- Nov. 03,
  2023, pp. 726--732.

\bibitem{chen2022unsupervised}
H.~Chen, N.~Yokoya, C.~Wu, and B.~Du, ``Unsupervised multimodal change
  detection based on structural relationship graph representation learning,''
  \emph{IEEE Transactions on Geoscience and Remote Sensing}, vol.~60, pp.
  1--18, Dec. 2022.

\bibitem{qu2021dual}
J.~Qu, Y.~Xu, W.~Dong, Y.~Li, and Q.~Du, ``Dual-branch difference amplification
  graph convolutional network for hyperspectral image change detection,''
  \emph{IEEE Transactions on Geoscience and Remote Sensing}, vol.~60, pp.
  1--12, Jan. 2022.

\bibitem{velivckovic2017graph}
P.~Veličković, G.~Cucurull, A.~Casanova, A.~Romero, P.~Liò, and Y.~Bengio,
  ``Graph attention networks,'' in \emph{Proc. International Conference on
  Learning Representations (ICLR)}, Vancouver, BC, Canada, Apr. 30-May 3, 2018.

\bibitem{wang2022csdbf}
X.~Wang, K.~Zhao, X.~Zhao, and S.~Li, ``{CSDBF}: Dual-branch framework based on
  temporal–spatial joint graph attention with complement strategy for
  hyperspectral image change detection,'' \emph{IEEE Transactions on Geoscience
  and Remote Sensing}, vol.~60, pp. 1--18, Oct. 2022.

\bibitem{zhang2023multi}
Y.~Zhang, R.~Miao, Y.~Dong, and B.~Du, ``Multiorder graph convolutional network
  with channel attention for hyperspectral change detection,'' \emph{IEEE
  Journal of Selected Topics in Applied Earth Observations and Remote Sensing},
  vol.~17, pp. 1523--1534, Jan. 2024.

\bibitem{dong2023local}
W.~Dong, Y.~Yang, J.~Qu, S.~Xiao, and Y.~Li, ``Local information-enhanced
  graph-transformer for hyperspectral image change detection with limited
  training samples,'' \emph{IEEE Transactions on Geoscience and Remote
  Sensing}, vol.~61, pp. 1--14, Apr. 2023.

\bibitem{lin2023hyperspectral}
T.-H. Lin and C.-H. Lin, ``Hyperspectral change detection using semi-supervised
  graph neural network and convex deep learning,'' \emph{IEEE Transactions on
  Geoscience and Remote Sensing}, vol.~61, pp. 1--18, Jun. 2023.

\bibitem{lin2021admm}
C.-H. Lin, Y.-C. Lin, and P.-W. Tang, ``{ADMM}-{ADAM}: A new inverse imaging
  framework blending the advantages of convex optimization and deep learning,''
  \emph{IEEE Transactions on Geoscience and Remote Sensing}, vol.~60, pp.
  1--16, Sep. 2022.

\bibitem{sagingalieva2023hybrid}
A.~Sagingalieva, M.~Kordzanganeh, N.~Kenbayev, D.~Kosichkina, T.~Tomashuk, and
  A.~Melnikov, ``Hybrid quantum neural network for drug response prediction,''
  \emph{Cancers}, vol.~15, no.~10, May 2023.

\bibitem{qu2023iomt}
Z.~Qu, W.~Shi, B.~Liu, D.~Gupta, and P.~Tiwari, ``Io{MT}-based smart healthcare
  detection system driven by quantum blockchain and quantum neural network,''
  \emph{IEEE Journal of Biomedical and Health Informatics}, vol.~28, no.~6, pp.
  3317--3328, Jun. 2024.

\bibitem{fan2023hybrid}
F.~Fan, Y.~Shi, T.~Guggemos, and X.~X. Zhu, ``Hybrid quantum-classical
  convolutional neural network model for image classification,'' \emph{IEEE
  Transactions on Neural Networks and Learning Systems}, pp. 1--15, 2023.

\bibitem{marvian2022restrictions}
I.~Marvian, ``Restrictions on realizable unitary operations imposed by symmetry
  and locality,'' \emph{Nature Physics}, vol.~18, no.~3, pp. 283--289, Jan.
  2022.

\bibitem{lin2023hyperqueen}
C.-H. Lin and Y.-Y. Chen, ``Hyper{QUEEN}: Hyperspectral quantum deep network
  for image restoration,'' \emph{IEEE Transactions on Geoscience and Remote
  Sensing}, vol.~61, pp. 1--20, May 2023.

\bibitem{hsu2024hyperqueen}
S.-M. Hsu, T.-H. Lin, and C.-H. Lin, ``{HyperQUEEN-MF}: Hyperspectral quantum
  deep network with multi-scale feature fusion for quantum image
  super-resolution,'' in \emph{Proc. 2024 IEEE 13rd Sensor Array and
  Multichannel Signal Processing Workshop (SAM)}, Corvallis, OR, USA, Jul.
  08-11, 2024, pp. 1--5.

\bibitem{lin2024prime}
C.-H. Lin and J.-T. Lin, ``{PRIME}: Blind multispectral unmixing using virtual
  quantum prism and convex geometry,'' \emph{arXiv preprint arXiv:2407.15358},
  2024.

\bibitem{he2016deep}
K.~He, X.~Zhang, S.~Ren, and J.~Sun, ``Deep residual learning for image
  recognition,'' in \emph{Proc. of the IEEE Conference on Computer Vision and
  Pattern Recognition (CVPR)}, Las Vegas, NV, USA, Jun. 26 - Jul. 1, 2016, pp.
  770--778.

\bibitem{xanthopoulos2013linear}
P.~Xanthopoulos, P.~M. Pardalos, and T.~B. Trafalis, ``Linear discriminant
  analysis,'' in \emph{Robust Data Mining}.\hskip 1em plus 0.5em minus
  0.4em\relax New York, NY: Springer New York, 2013, pp. 27--33.

\bibitem{achanta2012slic}
R.~Achanta, A.~Shaji, K.~Smith, A.~Lucchi, P.~Fua, and S.~Süsstrunk, ``{SLIC}
  superpixels compared to state-of-the-art superpixel methods,'' \emph{IEEE
  Transactions on Pattern Analysis and Machine Intelligence}, vol.~34, no.~11,
  pp. 2274--2282, Nov. 2012.

\bibitem{Qunitary}
M.~A. Nielsen and I.~L. Chuang, \emph{Quantum Computation and Quantum
  Information}.\hskip 1em plus 0.5em minus 0.4em\relax Cambridge University
  Press, Cambridge, 2010.

\bibitem{weigold2021encoding}
M.~Weigold, J.~Barzen, F.~Leymann, and M.~Salm, ``Encoding patterns for quantum
  algorithms,'' \emph{IET Quantum Communication}, vol.~2, no.~4, pp. 141--152,
  Dec. 2021.

\bibitem{senokosov2024quantum}
A.~Senokosov, A.~Sedykh, A.~Sagingalieva, B.~Kyriacou, and A.~Melnikov,
  ``Quantum machine learning for image classification,'' \emph{Machine
  Learning: Science and Technology}, vol.~5, no.~1, p. 015040, Mar. 2024.

\bibitem{liu2017band}
S.~Liu, Q.~Du, X.~Tong, A.~Samat, H.~Pan, and X.~Ma, ``Band selection-based
  dimensionality reduction for change detection in multi-temporal hyperspectral
  images,'' \emph{Remote Sensing}, vol.~9, no.~10, Sep. 2017.

\bibitem{liu2019unsupervised}
S.~Liu, Q.~Du, X.~Tong, A.~Samat, and L.~Bruzzone, ``Unsupervised change
  detection in multispectral remote sensing images via spectral-spatial band
  expansion,'' \emph{IEEE Journal of Selected Topics in Applied Earth
  Observations and Remote Sensing}, vol.~12, no.~9, pp. 3578--3587, Sep. 2019.

\bibitem{guo2021change}
Q.~Guo, J.~Zhang, C.~Zhong, and Y.~Zhang, ``Change detection for hyperspectral
  images via convolutional sparse analysis and temporal spectral unmixing,''
  \emph{IEEE Journal of Selected Topics in Applied Earth Observations and
  Remote Sensing}, vol.~14, pp. 4417--4426, Apr. 2021.

\bibitem{kingma2014adam}
D.~P. Kingma and J.~Ba, ``Adam: {A} method for stochastic optimization,'' in
  \emph{Proc. International Conference on Learning Representations (ICLR)},
  Y.~Bengio and Y.~LeCun, Eds., San Diego, CA, USA, May 7-9, 2015.

\end{thebibliography}

\begin{IEEEbiography}[{\resizebox{0.9in}{!}{\includegraphics[width=1in,height=1.25in,clip,keepaspectratio]{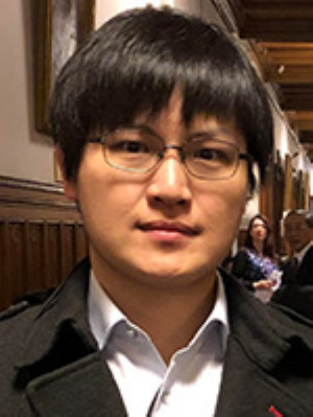}}}]
	{\bf Chia-Hsiang Lin}
	(S'10-M'18)
received the B.S. degree in electrical engineering and the Ph.D. degree in communications engineering from National Tsing Hua University (NTHU), Taiwan, in 2010 and 2016, respectively.
From 2015 to 2016, he was a Visiting Student of Virginia Tech,
Arlington, VA, USA.

He is currently an Associate Professor with the Department of Electrical Engineering, and also with 
the Miin Wu School of Computing,
National Cheng Kung University (NCKU), Taiwan.
Before joining NCKU, he held research positions with The Chinese University of Hong Kong, HK (2014 and 2017), 
NTHU (2016-2017), 
and the University of Lisbon (ULisboa), Lisbon, Portugal (2017-2018).
He was an Assistant Professor with the Center for Space and Remote Sensing Research, National Central University, Taiwan, in 2018, and a Visiting Professor with ULisboa, in 2019.
His research interests include network science, 
quantum computing,
convex geometry and optimization, blind signal processing, and imaging science.

Dr. Lin received the Emerging Young Scholar Award from National Science and Technology Council (NSTC), in 2023,
the Future Technology Award from NSTC, in 2022,
the Outstanding Youth Electrical Engineer Award from The Chinese Institute of Electrical Engineering (CIEE), in 2022,
the Best Young Professional Member Award from IEEE Tainan Section, in 2021,
the Prize Paper Award from IEEE Geoscience and Remote Sensing Society (GRS-S), in 2020, 
the Top Performance Award from Social Media Prediction Challenge at ACM Multimedia, in 2020,
and The 3rd Place from AIM Real World Super-Resolution Challenge at IEEE International Conference on Computer Vision (ICCV), in 2019. 
He received the Ministry of Science and Technology (MOST) Young Scholar Fellowship, together with the EINSTEIN Grant Award, from 2018 to 2023.
In 2016, he was a recipient of the Outstanding Doctoral Dissertation Award from the Chinese Image Processing and Pattern Recognition Society and the Best Doctoral Dissertation Award from the IEEE GRS-S.
\end{IEEEbiography}

\begin{IEEEbiography}[{\resizebox{1in}{!}{\includegraphics[width=1in,height=1.25in,clip,keepaspectratio]{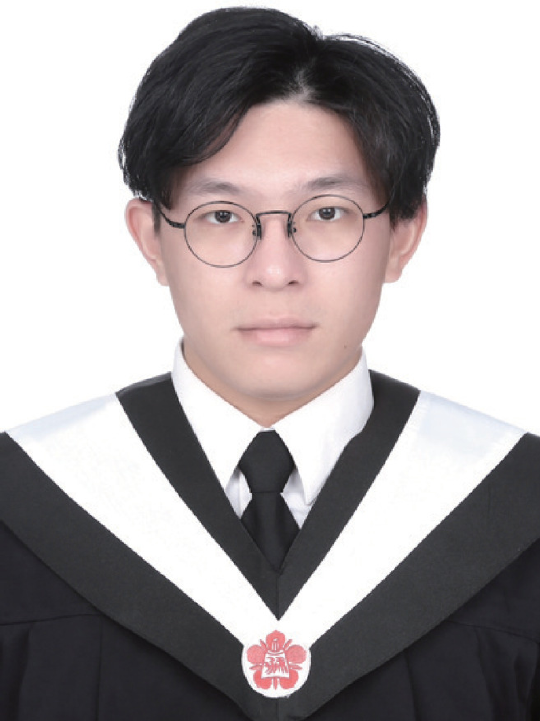}}}]
	{\bf Tzu-Hsuan Lin}
	(S'20)
	received his B.S. degree from the Department of Electrical Engineering, National Cheng Kung University (NCKU), Taiwan, in 2019.
 
 He is currently a Visiting Student with INRIA (Université Grenoble Alpes), Grenoble, France, and also a Ph.D. student with Intelligent Hyperspectral Computing Laboratory, Institute of Computer and Communication Engineering, NCKU, Taiwan. 
	His research interests include deep learning, convex optimization, change detection, compressed sensing, and hyperspectral imaging. 

Mr. Lin was a recipient of the Outstanding Paper Award from the Chinese Image Processing and Pattern Recognition Society (IPPR) Conference on Computer Vision, Graphics, and Image Processing (CVGIP), in 2022.
He has been selected as a recipient of the Ph.D. Students Study Abroad Program from the National Science and Technology Council (NSTC), Taiwan, in 2024.
\end{IEEEbiography}

\begin{IEEEbiography}[{\resizebox{1in}{!}{\includegraphics[width=1in,height=1.25in,clip,keepaspectratio]{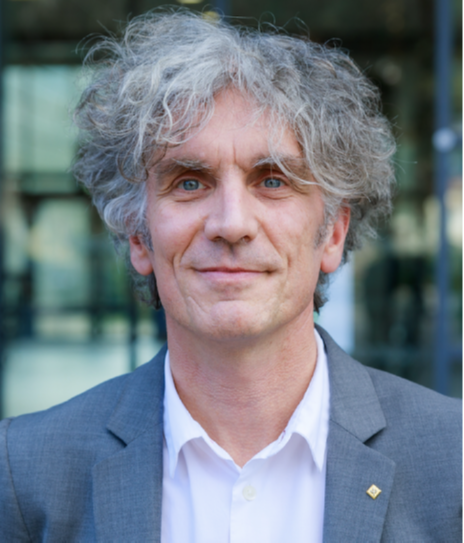}}}]
    {\bf Jocelyn Chanussot}
    (IEEE Fellow)
    received the M.Sc. degree in electrical engineering from the Grenoble Institute of Technology (Grenoble INP), Grenoble, France, in 1995, and the Ph.D. degree from the Université de Savoie, Annecy, France, in 1998.

    From 1999 to 2023, he was with Grenoble INP, where he was a Professor of signal and image processing.
    He has been a Visiting Scholar at Stanford University, Stanford, CA, USA; KTH, Stockholm, Sweden; and the National University of Singapore, Singapore.
    Since 2013, he has been an Adjunct Professor at the University of Iceland, Reykjavík, Iceland.
    From 2015 to 2017, he was a Visiting Professor at the University of California at Los Angeles (UCLA), Los Angeles, CA, USA.
    He is currently a Research Director with INRIA, Grenoble.
    He holds the AXA chair in remote sensing and is an Adjunct Professor with the Aerospace Information Research Institute, Chinese Academy of Sciences, Beijing, China.
    His research interests include image analysis, hyperspectral remote sensing, data fusion, machine learning, and artificial intelligence.
    
    Dr. Chanussot is a fellow of ELLIS and AAIA, a member of the Institut Universitaire de France from 2012 to 2017, and has been a Highly Cited Researcher (Clarivate Analytics/Thomson Reuters) since 2018.
    He was the Founding President of the IEEE Geoscience and Remote Sensing French Chapter from 2007 to 2010, which received the 2010 IEEE GRSS Chapter Excellence Award.
    He was the Vice-President of the IEEE Geoscience and Remote Sensing Society, in charge of meetings and symposia, from 2017 to 2019.
    He was the Editor-in-Chief of IEEE Journal of Selected Topics in Applied Earth Observations and Remote Sensing from 2011 to 2015.
    In 2014, he served as a Guest Editor for IEEE Signal Processing Magazine.
    He is an Associate Editor of IEEE Transactions on Geoscience and Remote Sensing, IEEE Transactions on Image Processing, and Proceedings of the IEEE.
\end{IEEEbiography}

\end{document}